%% file: arxiv2.tex
\documentclass{article} % For LaTeX2e
\usepackage{iclr2026_conference,times}

\input{math_commands.tex}

\usepackage{hyperref}
\usepackage{url}
\usepackage{graphicx}
\usepackage{subcaption}
\usepackage{pifont}
\usepackage[table]{xcolor}
\usepackage{tikz}
\newcommand*\circledtext[1]{%
  \tikz[baseline=(char.base)]{
    \node[shape=circle,draw,inner sep=0.5pt, font=\footnotesize] (char) {#1};}}

\title{Linking Process to Outcome: Conditional Reward Modeling for LLM Reasoning}

\author{
Zheng Zhang\textsuperscript{1,2},
Ziwei Shan\textsuperscript{1},
Kaitao Song\textsuperscript{3},
Yexin Li\textsuperscript{\,2}\footnotemark[1] \,,
Kan Ren\textsuperscript{1}\thanks{Corresponding Author.} \\
\textsuperscript{1}School of Information Science and Technology, ShanghaiTech University \\ 
\textsuperscript{2}State Key Laboratory of General Artificial Intelligence, BIGAI 
\\
\textsuperscript{3}Independent Researcher\\
\texttt{\{zhangzheng2024,shanzw2022,renkan\}@shanghaitech.edu.cn} \\
\texttt{liyexin@bigai.ai}
\\
\texttt{stillkeeptry@outlook.com}
}

\usepackage{xspace}
\newcommand{\method}{\textsc{CRM}\xspace}
\usepackage{booktabs}
\usepackage{multirow}
\usepackage{makecell}
\usepackage{wrapfig} 
\usepackage{minted}
\usepackage{caption}

\usepackage[most]{tcolorbox}

\iclrfinalcopy % Uncomment for camera-ready version, but NOT for submission.
\begin{document}

\maketitle

\begin{abstract}
Process Reward Models (PRMs) have emerged as a promising approach to enhance the reasoning capabilities of large language models (LLMs) by guiding their step-by-step reasoning toward a final answer.
However, existing PRMs either treat each reasoning step in isolation, failing to capture inter-step dependencies, or struggle to align process rewards with the final outcome. 
Consequently, the reward signal fails to respect temporal causality in sequential reasoning and faces ambiguous credit assignment.
These limitations make downstream models vulnerable to reward hacking and lead to suboptimal performance.
In this work, we propose Conditional Reward Modeling (\method) that frames LLM reasoning as a temporal process leading to a correct answer.
The reward of each reasoning step is not only conditioned on the preceding steps but also explicitly linked to the final outcome of the reasoning trajectory. By enforcing conditional probability rules, our design captures the causal relationships among reasoning steps, with the link to the outcome allowing precise attribution of each intermediate step, thereby resolving credit assignment ambiguity.
Further, through this consistent probabilistic modeling, the rewards produced by \method enable more reliable cross-sample comparison.
Experiments across Best-of-N sampling, beam search and reinforcement learning demonstrate that \method consistently outperforms existing reward models, offering a principled framework for enhancing LLM reasoning.
In particular, \method is more robust to reward hacking and delivers stable downstream improvements without relying on verifiable rewards derived from ground truth.
The project website is at \url{https://foundation-model-research.github.io/CRM}.
\end{abstract}

\section{Introduction}
Recent advances in enhancing reasoning abilities have greatly improved the performance of large language models (LLMs) \citep{snell2024scaling, jaech2024openai}, where models derive final answers through explicit step-by-step reasoning. 
Beyond prompt-based approaches \citep{wei2022chain, diao2023active, fan2025improving}, state-of-the-art systems like DeepSeek-R1 \citep{guo2025deepseek} have further advanced reasoning capacity through reinforcement learning (RL) with verifiable rewards.

However, verifiable rewards rely on checking model outputs against ground-truth labels, whose acquisition is costly and difficult to scale for general reasoning improvements. 
Reward models provide a promising alternative by extrapolating reward signals across general LLM reasoning processes. 
Broadly, these models fall into two categories: outcome reward models (ORMs) \citep{cobbe2021training,yu2023ovm}, which provide feedback at the final step;
and process reward models (PRMs) \citep{wang2023math,li2024process,yuan2024free}, which provide finer-grained rewards at the level of individual reasoning steps or even tokens.

\begin{figure}[t]
  \centering
  \begin{subfigure}{0.63\linewidth}
    \centering
    \includegraphics[width=\linewidth]{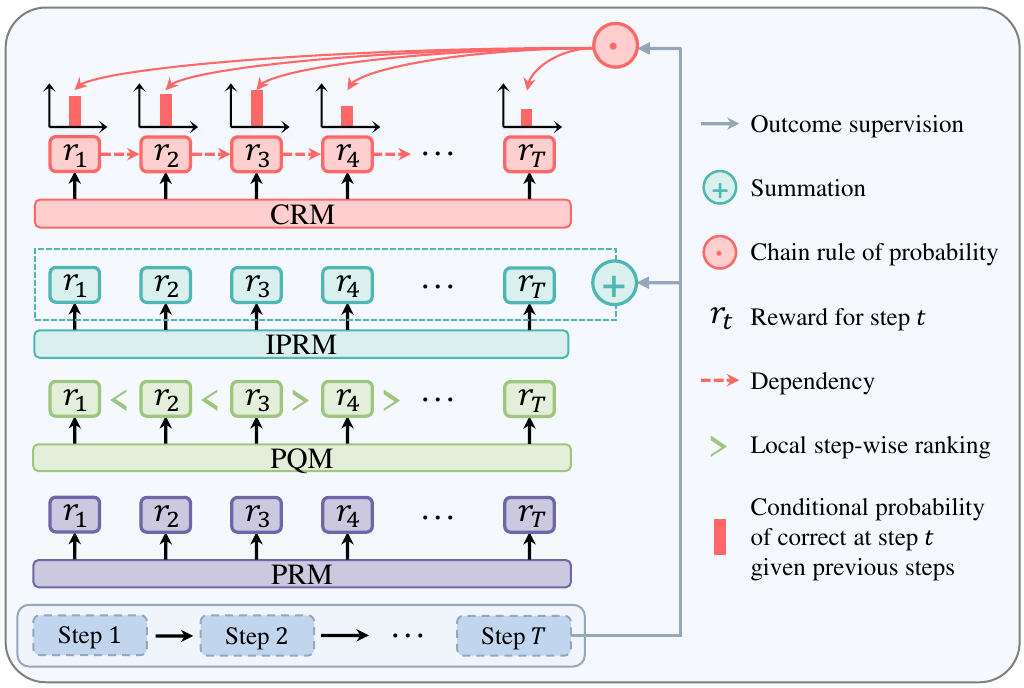}
    \caption{}
    \label{fig:introduction_a}
  \end{subfigure}
  \hfill
  \begin{subfigure}{0.36\linewidth}
    \centering
    \includegraphics[width=\linewidth]{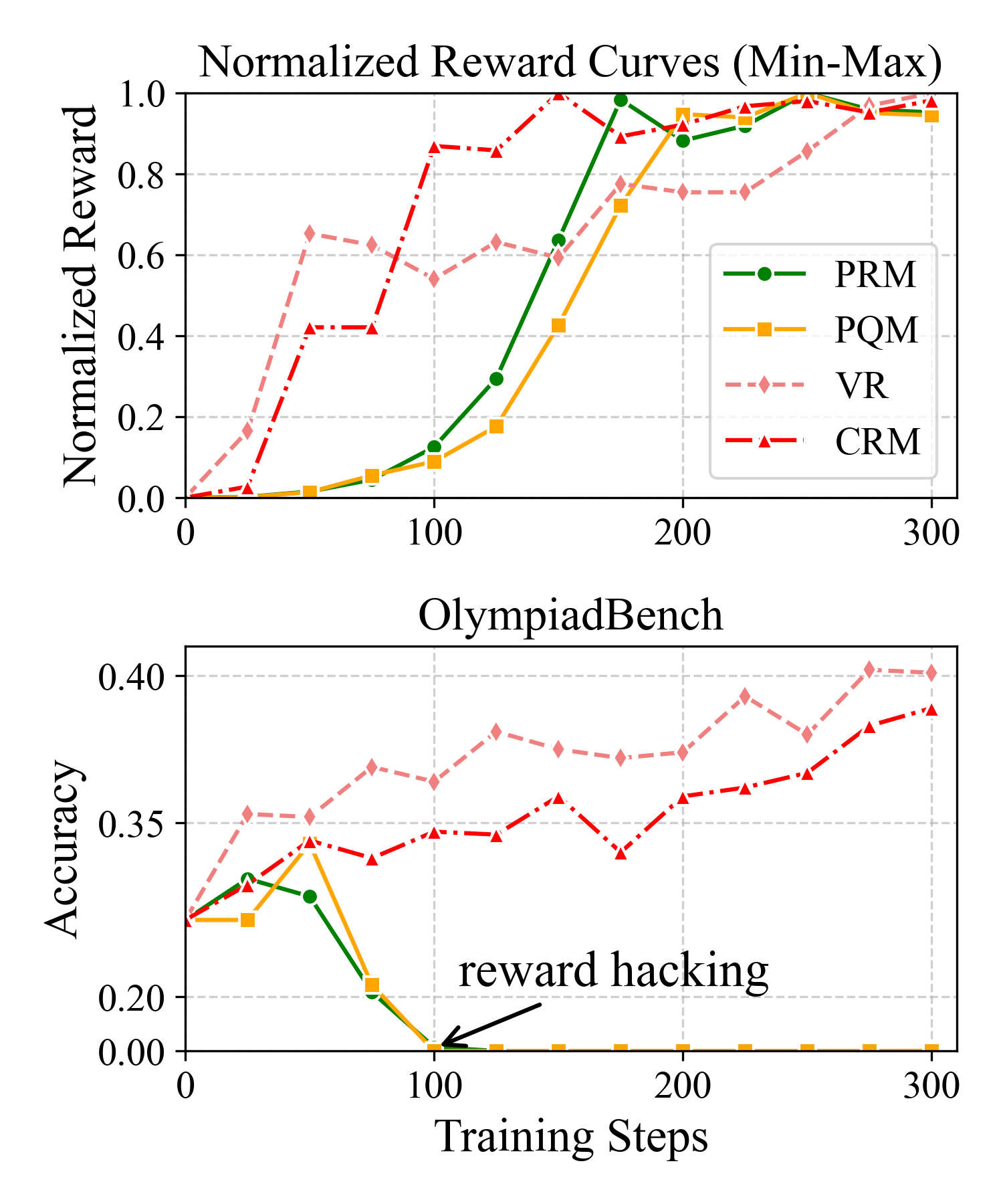}
    \caption{}
    \label{fig:introduction_b}
  \end{subfigure}
  \vspace{-15pt}
  \caption{
  (a) Comparison of reward modeling paradigms. Our \method explicitly conditions each step reward on the previous reasoning steps and aligns it with the final outcome. 
  (b) RL training with different reward models. 
  Our \method is more robust to reward hacking and achieves performance on par with training using verifier rewards (VR).
  }
  \label{fig:introduction}
  \vspace{-10pt}
\end{figure}

Despite offering nuanced signals for reasoning processes, existing PRMs face several limitations. 
(i) \textit{Isolated step modeling}: As shown in Figure~\ref{fig:introduction_a}, most PRMs \citep{lightman2023let,wang2023math,luo2024improve,shao2024deepseekmath} assess each reasoning step in isolation, neglecting the intrinsic sequential dependencies of reasoning.
(ii) \textit{Limited outcome awareness}: While some methods \citep{yu2023ovm,li2024process,yuan2024free} attempt to mitigate isolated step rewards, they often fail to effectively link step-wise rewards with the outcome.
For example, PQM \citep{li2024process}, illustrated in Figure~\ref{fig:introduction_a}, relies on relative comparisons between neighboring steps (e.g., greater or smaller reward), but lacks explicit modeling of the final outcome, which may result in biased process rewards.
IPRM \citep{yuan2024free} parameterizes the outcome reward as the logarithmic sum of process rewards, yet it fails to capture how a specific step relates to the final outcome and lacks nuanced modeling of inter-step dependencies.
This leads to ambiguous credit assignment from the final outcome back to the intermediate reasoning steps.
As a result, existing methods are prone to reward hacking, where rewards continue to increase while the actual task accuracy declines, which we have observed in our experiments and has been shown in Figure~\ref{fig:introduction_b}.

To address these limitations, we propose Conditional Reward Modeling (\method), which frames reasoning as \textit{a temporal process through which an LLM progressively approaches the correct final answer}. 
In our formulation, the reward signal is modeled as the probabilistic evolution of deriving the correct outcome conditioned on the existing reasoning steps.
At each step, the process reward is treated as a conditional probability dependent on all the preceding steps, thereby capturing causal structure inherent in sequential reasoning.
Furthermore, by explicitly linking each process reward to the final outcome via the conditional probability chain rule, \method enables precise attribution of the final result to individual reasoning steps, effectively resolving the credit assignment ambiguity prevalent in prior work.
An additional advantage is that the probabilistically consistent formulation of process reward signals across reasoning trajectories facilitates cross-sample comparison, which significantly benefits downstream tasks such as Best-of-N sampling, beam search, and RL optimization.
As shown in Figure~\ref{fig:introduction_a}, \method jointly models inter-step dependencies and the relationship between intermediate rewards and the final outcome.

Our contributions are three-fold as follows.
(i) \textit{Conditional reward modeling framework}: 
We introduce \method, which defines each step’s reward as a conditional probability dependent on all preceding steps, thereby capturing inter-step dependencies.  
(ii) \textit{Precise credit assignment}: By linking process rewards to the final outcome, \method resolves the ambiguity of credit assignment in existing PRMs.
(iii) \textit{Practical effectiveness and robustness}: 
\method enhances cross-sample comparison and improves various downstream tasks. 
Experiments demonstrate that our \method achieves superior performance, remains robust to reward hacking and delivers stable reasoning improvements without reliance on a verifier that uses ground-truth labels.

\section{Related Work}
\paragraph{Enhancing the Reasoning Ability of LLMs.}
To enhance LLM reasoning, prior studies have explored test-time search (e.g., Best-of-N, beam search) with reward models to select promising solutions \citep{xie2023self, zhang2024generative, snell2024scaling}, and post-training via RL \citep{jaech2024openai,guo2025deepseek}.
Recent work \citep{zeng2503simplerl, guo2025deepseek} has investigated RL with verifiable rewards for reasoning tasks.
However, these approaches heavily rely on ground-truth and thus do not easily scale.
Other studies \citep{cui2025process,cheng2025stop} integrate reward models into RL to provide dense feedback, but reward hacking remains a major concern \citep{gao2024designing}.

\paragraph{Process Reward Model.}
Most existing studies \citep{lightman2023let, wang2023math,luo2024improve, shao2024deepseekmath} treat PRM as step-level classification, which treats steps in isolation and ignores inter-step dependencies.
Recent studies \citep{lu2024autopsv, li2024process, yuan2024free} attempt to move beyond classification-based PRMs. 
\citet{li2024process} reframes PRM as a Q-value ranking problem to model the relationships between steps, but overlooks explicit modeling of the final outcome, resulting in a gap between process rewards and the final outcome.
\citet{yuan2024free} instead proposes a parameterized formulation of outcome, leveraging outcome labels to train the PRM, but the relationship between intermediate step rewards and the final outcome remains unclear, and the lack of step dependency modeling leads to ambiguous credit assignment.
Another line of work \citep{chen2025rm, guo2025reward, she-etal-2025-r, zhao2025genprm} integrates reasoning capabilities into reward modeling, such as R-PRM \citep{she-etal-2025-r} and GenPRM \citep{zhao2025genprm}, which evaluate reasoning processes with detailed thinking. However, these approaches often involve complex training procedures and incur high inference costs.

\section{Methodology}
\subsection{Task Description}
Given a question $x$, LLM $\pi$ generates a response $y=(a_1, a_2,\ldots, a_T)$ where $a_t$ denotes the $t$-th reasoning step and $T$ is the total step number.
Let $a_{\leq t}$ denote the sequence of the first $t$ steps.
Each response is assigned a binary label $l \in \{0, 1\}$ indicating whether the final answer derived from it is correct.

We model multi-step reasoning as a finite-horizon Markov Decision Process (MDP) $\mathcal{M} = ( \mathcal{S}, \mathcal{A}, \mathcal{P}, r, \gamma )$ driven by an autoregressive LLM $\pi$ acting as the policy.
$\mathcal{S}$ is the state space, $\mathcal{A}$ is the action space, $\mathcal{P}$ represents transition dynamics, $r: \mathcal{S} \times \mathcal{A} \to \mathbb{R}$ is the reward function, and $\gamma \in [0, 1]$ is the discount factor.
At reasoning step $t$, the state is defined as $s_t = (x, a_{\leq t-1})$, which contains the question $x$ and the sequence of previously generated reasoning steps $a_{\leq t-1}$.
The action $a_t$ is a reasoning step generated based on the state $s_{t}$.
The state transition is deterministic, as the next state is uniquely determined by concatenating the previous sequence with the current output $a_t$.
$r_t$ denotes the reward provided by the environment or generated by the reward model for action $a_t$.
\subsection{\method Modeling}
\label{sec:Modeling}
To remedy isolated step modeling and limited outcome awareness in existing PRMs, we model LLM reasoning as a temporal process in which the probability of reaching the correct answer evolves with step $t$.
However, in practice, it is difficult to directly quantify the extent to which reasoning is approaching the correct answer. 
Instead, we choose to model the complementary event: the reasoning process entering a wrong state, which implies that the reasoning trajectory can no longer yield the correct answer. 
Formally, we define $z$ as the index of the first step at which the reasoning process enters such a wrong state, with $z \geq 1$.
If no wrong state occurs throughout the trajectory, then $z > T$, meaning the reasoning is correct and the final answer is correct ($l=1$). Conversely, if the final answer is incorrect ($l=0$), then the trajectory has entered a wrong state during reasoning, with $z \leq T$.
Let $p(z; x, a_{\leq z})$ denote the probability mass function of a wrong state occurring at step $z$. 
Accordingly, the probability that a wrong state has already occurred at or before step $t$ is
\begin{equation}
W(t; x, a_{\leq t}) = \Pr(z \leq t) = \sum_{z=1}^t p(z; x, a_{\leq z})
\end{equation}
The probability of maintaining correct reasoning up to step $t$ is its complement,
\begin{equation}
S(t; x, a_{\leq t}) = \Pr(z > t) = 1 - W(t; x, a_{\leq t}) = \sum_{z = t+1}^\infty p(z; x, a_{\leq t})
\end{equation}
The discrete probability mass function of the wrong state occurring at step $t$ is
\begin{equation}
p(t; x, a_{\leq t}) = \Pr(z = t) = W(t; x, a_{\leq t}) - W(t-1; x, a_{\leq t-1})
% = S(t-1; x, a_{\leq t-1}) - S(t; x, a_{\leq t})
\end{equation}
For simplicity, we use $p(t)$, $W(t)$, $S(t)$ to denote $p(t; x, a_{\leq t})$, $W(t; x, a_{\leq t})$ and $S(t; x, a_{\leq t})$.

A reasoning trajectory is inherently causal: the correctness of step $t$ logically depends on all preceding $(t\!-\!1)$ steps. 
Building on this view, we adopt a conditional probability perspective and derive $h(t)$,
the probability that step $t$ enters a wrong state given that all previous $(t\!-\!1)$ steps were correct.
\begin{equation}
h(t) = \Pr(z = t | z \geq t) = \frac{\Pr(z = t)}{\Pr(z \geq t)} = \frac{p(t)}{S(t-1)}
\end{equation}
Naturally, $1\!-\!h(t)$ corresponds to the complementary event: the probability that the current step is correct given that all previous $(t-1)$ steps were correct.
This formulation explicitly captures step dependencies through conditional probabilities, addressing the limitations of prior studies \citep{lightman2023let, wang2023math,luo2024improve, shao2024deepseekmath} that overlook inter-step relations.

To explicitly link intermediate steps to the final outcome, we first apply the chain rule of probability to establish the relationships among $S(t)$, $W(t)$, $p(t)$, and $h(t)$.
\begin{multline}
S(t) = \Pr(z > t) = \Pr(z \ne 1, z \ne 2,..., z \ne t) 
= \Pr(z \ne 1) \cdot \Pr(z \ne 2 | z \ne 1) \cdots \\ \cdots \Pr(z \ne t | z \ne 1, z \ne 2..., z \ne t-1) 
= \prod_{k=1}^t \left[1 - \Pr (z = k \mid z \geq k ) \right] = \prod_{k=1}^t (1 - h(k) )
\label{eq:S_t_decompose}
\end{multline}
\begin{equation}
W(t) = \Pr(z \leq t) = 1- S(t) = 1 - \prod_{k=1}^t (1 - h(k) )
\end{equation}
\begin{equation}
p(t) = \Pr(z = t) = h(t) \prod_{k=1}^{t-1} (1-h(k))
\end{equation}
For the final step of the reasoning trajectory $T$, we have $S(T) = \prod_{t=1}^T (1 - h(t))$ from Eq.~\ref{eq:S_t_decompose}, where $S(T)$ reflects the probability that the reasoning process reaches the correct final answer.
We next seek a dense, step-wise reward signal aligned with this outcome probability.

We apply Potential-Based Reward Shaping (PBRS) \citep{ng1999policy} to the task of multi-step reasoning.
PBRS proposes that a dense reward function $R'$ can be constructed from an original sparse reward function $R$ by adding a shaping term derived from a potential function $\Phi(s)$ defined over the state space. For a transition from state $s_t$ to $s_{t+1}$, the shaped reward is given by:
\begin{equation}
    R'(s_t, a_t, s_{t+1}) = R(s_t, a_t) + \gamma \Phi(s_{t+1}) - \Phi(s_t)
    \label{eq:pbrs}
\end{equation}
PBRS guarantees that any policy optimal under the shaped reward $R'(s_t, a_t, s_{t+1})$ is also optimal under the original reward $R(s_t, a_t)$. 
Thus, using the shaped reward guides the learning process more effectively without changing the fundamental goal of the task.
The key to applying PBRS lies in selecting an appropriate potential function $\Phi(s_t)$
which is effective when it encodes goal-directed progress \citep{JMLR:v25:24-0040, muller2025improving, mindom2021assessing, wiewiora2003principled}.
We define the potential function as $\log S(t)$, which denotes the log-likelihood of eventually reaching a correct answer from the current state, thereby explicitly capturing the progress toward the final correct answer.
Consequently, substituting this definition of $\log S(t)$ into the potential function yields the following explicit form:
\begin{equation}
    \Phi(s_t) \equiv \log S(t) = \log \left( \prod_{k=1}^t (1 - h(k)) \right) = \sum_{k=1}^t \log(1 - h(k))
\end{equation}
We now apply the general PBRS formula (Eq. \ref{eq:pbrs}) to derive the specific process reward $r_t$, which corresponds to the transition from state $s_{t-1}$ to $s_t$.
\begin{equation}
    r_t \equiv R'(s_{t-1}, a_{t-1}, s_t) = R(s_{t-1}, a_{t-1}) + \gamma \Phi(s_t) - \Phi(s_{t-1}) 
    = \log(1 - h(t))
\end{equation}
where original reward $R = 0$ for all intermediate steps, and the discount factor is fixed at $\gamma = 1$.

The above derivation yields a dense and accurate process reward, $r_t = \log(1 - h(t))$, which serves as a credit assignment scheme grounded in our modeling.
The probability that the reasoning process reaches the correct final answer, $S(T)$, satisfies $S(T) = \prod_{t=1}^T (1 - h(t)) = \prod_{t=1}^T e^{r_t}$.
Through this decomposition, we explicitly link process rewards to the outcome, addressing the limitation of prior work \citep{yu2023ovm, li2024process, yuan2024free} that lacks explicit modeling of the relationship between process and outcome.

\subsection{Training of Conditional Reward Model}
Let $\mathcal{D}=\{(x_i, y_i, l_i, z_i) \}$ denote the training dataset, where $x_i$ is the $i$-th question, $y_i$ is the corresponding response (multi-step reasoning trajectory), $l_i$ indicates whether the final answer is correct, and $z_i$ is the index of the first wrong state.
We use a large language model $f_\phi$ with parameters $\phi$ to initialize the reward model.
Since both $S(T)$ and the process reward $r_t$ are functions of $h(t)$, we train the model to predict $h(t)$.
% depend on
At the $t$-th step of the reasoning trajectory, the model takes the question $x$ and the first $t$ reasoning steps as input, and predicts the $h(t) = f_{\phi}(x, a_{\leq t})$.

\begin{wrapfigure}[10]{r}{0.48\columnwidth}
  \vspace{-1.2\baselineskip}
  \centering
  \includegraphics[width=\linewidth]{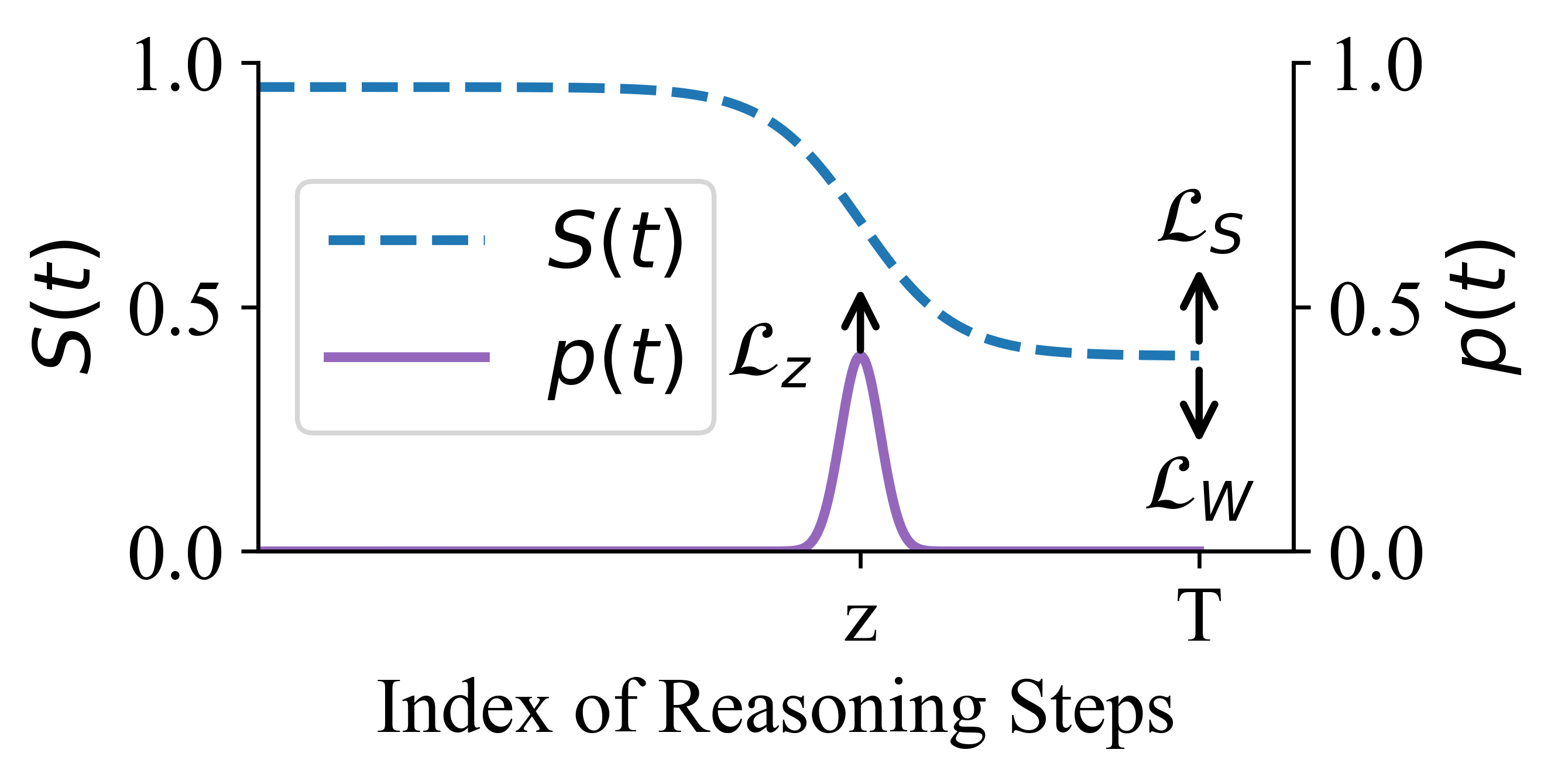}
  \vspace{-2.0\baselineskip}
  \caption{Effects of three loss terms.}
  \label{fig:visualization_losses}
\end{wrapfigure}

For samples where the reasoning process reaches the correct final answer ($l_i = 1$), we maximize $S(T)$ and the corresponding loss $\mathcal{L}_S$ is:
\begin{equation}
\mathcal{L}_S (x_i,y_i) =  -\log \Pr(z_i > T) = -\log S(T) = -\log \left[ \prod_{t=1}^T (1 - h(t) ) \right] 
% = - \sum_{t=1}^T \log \left(1 - h(t)\right)
\label{eq:L_S}
\end{equation}
For samples where the reasoning trajectory fails to reach the correct final answer ($l_i = 0$), we minimize $S(T)$.
Moreover, since the reasoning process enters a wrong state at step $z_i$, we encourage the model to identify this step by maximizing $p(z_i)$, the probability of a wrong state occurring exactly at step $z_i$.
The loss $\mathcal{L}_W$ and $\mathcal{L}_z$ for this case are as follows:
\begin{equation}
\mathcal{L}_W (x_i,y_i) =  -\log \Pr(z_i \leq T) = -\log (1-S(T)) = -\log \left[1 - \prod_{t=1}^T (1 - h(t) ) \right]
\label{eq:L_W}
\end{equation}
\begin{equation}
\mathcal{L}_z (x_i,y_i,z_i) =  -\log \Pr(z_i) = -\log p(z_i) = -\log \left[ h(z_i) \prod_{t=1}^{z_i-1} (1-h(t)) \right]
\label{eq:L_z}
\end{equation}
Figure \ref{fig:visualization_losses} illustrates the roles of the three loss terms.
The overall loss to be minimized is as follows:
\begin{equation}
\mathcal{L}
= \frac{1}{\vert \mathcal D\vert} \sum_{i=1}^{\vert \mathcal D\vert}
\Bigl[
  l_i\,\mathcal{L}_S(x_i,y_i)
  +
  (1 - l_i)\bigl(\mathcal{L}_W(x_i,y_i) 
    + \mathcal{L}_z(x_i,y_i,z_i)\bigr)
\Bigr]
\end{equation}
\label{sec:Model_Training}
This consistent probabilistic modeling and training ensure that \method is grounded in clear probabilistic semantics:
for example, the value of $S(t)$ at any step $t$ across different samples consistently carries the same probabilistic meaning, allowing them to be compared under the same scale.
By contrast, previous approaches \citep{li2024process, yuan2024free} struggle to achieve accurate cross-sample comparison (see Appendix \ref{appendix:cross-sample} for detailed analysis).

\vspace{-3pt}
\section{Experiments}
\vspace{-3pt}
\subsection{Experimental Settings}
\vspace{-3pt}
\textbf{Reward Model Training.}
We train the reward model on the Math-Shepherd dataset \citep{wang2023math}, which integrates questions from GSM8K \citep{cobbe2021training} and MATH \citep{hendrycks2021measuring} and provides responses annotated with step-level process labels.
Following prior research \citep{li2024process, guan2025rstar}, we augment the pre-trained model with a value head.
The reward models are trained using full parameter fine-tuning. 
Further details can be found in Appendix \ref{appendix:implementation_details}.

\textbf{Baselines and Evaluation.}
We compare our \method against representative baselines, including ORM, vanilla PRM \citep{wang2023math}, PQM \citep{li2024process} and IPRM \citep{yuan2024free}. 
For fairness, all baselines are re-implemented within the same pipeline, backbone,  and training data.
We then leverage the rewards provided by the reward model to (i) select an optimal response in Best-of-N (Section \ref{sec:BoN}), (ii) guide beam search (Section \ref{sec:Beam Search}), and (iii) optimize LLM reasoning through RL (Section \ref{sec:RL Optimization}).

\subsection{Best-of-N Sampling Experiments}
\label{sec:BoN}
Best-of-N sampling generates $N$ responses for a given question and selects the optimal one using a reward model, evaluating the model’s ability to identify correct samples at the trajectory level.
Our \method computes $S(T)$ as the trajectory-level score, and we evaluate it on two popular math reasoning datasets: GSM-Plus \citep{li2024gsm} and MATH500 \citep{lightman2023let}. 

\begin{table}[ht]
\caption{Best-of-N accuracy across models. 
\textbf{Bold} and \underline{underlined} values denote the top two results.}

\label{tab:BoN}
\centering
\renewcommand{\arraystretch}{1.2}
\resizebox{\linewidth}{!}{
\begin{tabular}{llccccc|ccccc}
\toprule
\multirow{2}*{\textbf{Models}} & \multirow{2}*{\textbf{Methods}} & \multicolumn{5}{c|}{\textbf{GSM-Plus}} & \multicolumn{5}{c}{\textbf{MATH500}} \\
& & @8 & @16 & @32 & @64 & @128 & @8 & @16 & @32 & @64 & @128 \\
\midrule
% & Major@N & 66.3 & 65.6 & 64.7 & 64.2 & 64.2 & 46.2 & 46.0 & 45.2 & 45.0 & 44.8 \\
% & Pass@N (Oracle) & 79.2 & 81.2 & 82.2 & 83.3 & 85.5 & 73.8 & 79.2 & 84.6 & 87.4 & 90.2 \\
& Major@N & 66.3 & 65.6 & 64.7 & 64.2 & 64.2 & 46.2 & 46.0 & 45.2 & 45.0 & 44.8 \\
& Pass@N (Oracle) & 79.2 & 81.2 & 82.2 & 83.3 & 85.5 & 73.8 & 79.2 & 84.6 & 87.4 & 90.2 \\
\midrule
\multirow{5}{*}{\shortstack{Qwen2.5-\\3B-Instruct}}
    & ORM & 66.8 & 67.2 & 66.4 & 65.7 & 65.7 & 51.6 & 51.4 & 51.8 & 49.0 & 49.2 \\
    & PRM & 67.6 & 67.9 & 67.7 & 66.9 & 66.7 & \textbf{54.2} & \underline{55.2} & \underline{55.2} & 54.2 & 54.6 \\
    & PQM & \textbf{68.5} & \textbf{69.2} & \textbf{68.5} & \underline{68.2} & \underline{68.0} & \underline{53.2} & 54.4 & 54.8 & \underline{54.8} & \underline{55.8} \\
    & IPRM & 65.5 & 66.2 & 66.8 & 66.5 & 66.2 & 52.4 & 52.0 & 52.0 & 52.2 & 53.0 \\
    \rowcolor{blue!10}\cellcolor{white} & \textbf{\method (ours)} & \underline{67.8} & \underline{68.6} & \underline{67.9} & \textbf{68.4} & \textbf{68.7} & 53.0 & \textbf{56.4} & \textbf{56.6} & \textbf{55.8} & \textbf{56.6} \\
\midrule
\multirow{5}{*}{\shortstack{LLaMA3.1-\\8B}}
    & ORM        & 66.9 & 67.4 & 67.1 & 67.2 & 66.6 & 47.4 & 46.4 & 44.6 & 45.2 & 45.6 \\
    & PRM        & \textbf{67.9} & \underline{68.2} & \underline{68.5} & \textbf{68.8} & \textbf{68.9} & 48.0 & 48.0 & \underline{49.8} & 49.0 & 47.6 \\
    & PQM        & 66.4 & 67.0 & 66.2 & 66.5 & 67.2 & \textbf{51.0} & \textbf{51.4} & 48.8 & \underline{49.0} & \underline{48.4} \\
    & IPRM& 65.1 & 64.3 & 63.9 & 63.5 & 63.7 & 48.4 & 45.8 & 44.2 & 46.0 & 45.8 \\
    \rowcolor{blue!10}\cellcolor{white} & \textbf{\method (ours)} & \underline{67.8} & \textbf{68.8} & \textbf{69.1} & \underline{68.6} & \underline{68.5} & \underline{49.4} & \underline{50.6} & \textbf{50.6} & \textbf{49.8} & \textbf{50.6} \\
\bottomrule
\end{tabular}
}
\end{table}

\paragraph{\method demonstrates stronger trajectory-level selection in Best-of-N sampling.}
\begin{wrapfigure}[10]{r}{0.4\columnwidth}
  \vspace{-1\baselineskip}
  \centering
  \includegraphics[width=\linewidth]{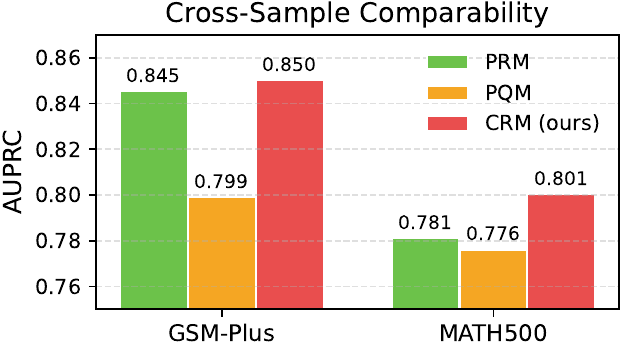}
  \vspace{-18pt}
  \caption{Cross-sample comparability.
  }
  \label{fig:BoN_auprc}
  \vspace{-40pt}
  % \vspace{-2\baselineskip}
\end{wrapfigure}
As shown in Table \ref{tab:BoN}, \method consistently ranks at or near the top across both datasets and model families.
For example, on MATH500 with Qwen2.5-3B-Instruct, \method reaches 56.6\% at $N$=32, surpassing the strongest baseline by +1.4 and remaining ahead across $N$=16, 64, and 128. 
By conditioning each step’s reward on all preceding steps, \method ensures that the trajectory score reflects the coherence of the entire reasoning chain rather than isolated fragments. 
This holistic formulation enables \method to assess logical consistency across steps and more reliably distinguish correct trajectories from superficial ones, leading to stronger trajectory-level selection.
A more in-depth analysis of the performance characteristics of BoN is provided in Appendix~\ref{appendix:BoN_analysis}.

\paragraph{\method shows superior cross-sample comparability.}
Best-of-N evaluation focuses on the model’s ability to identify the correct response among $N$ responses for the same question, where our \method already shows clear advantages.
To move beyond this setting and assess cross-question comparability, we mix all responses from different questions and adopt AUPRC (Area Under the Precision–Recall Curve) as the metric, which measures whether correct responses are concentrated among those with higher rewards.
A higher AUPRC indicates that when ranking all responses globally by reward score, correct responses are consistently placed toward the top.
As shown in Figure \ref{fig:BoN_auprc}, our \method outperforms the baselines, validating its superior cross-sample comparability.
This advantage arises from the consistent probabilistic modeling (see Section~\ref{sec:Model_Training}), which ensures that reward signals carry the same semantic meaning across reasoning trajectories, thereby making them directly comparable across samples.

\begin{table}[ht]
\centering

\caption{Beam Search accuracy on MATH500 and Gaokao2023.}

\resizebox{\textwidth}{!}{
    \begin{tabular}{ll|cccc|cccc}
        \toprule
        \multirow{2}*{\textbf{Models}} & \multirow{2}*{\textbf{Methods}} & \multicolumn{4}{c|}{\textbf{MATH500}}  & \multicolumn{4}{c}{\textbf{GAOKAO2023}} \\
        & 
        & $N=4$ & $N=8$ & $N=20$ & $N=100$
        & $N=4$ & $N=8$ & $N=20$ & $N=100$\\
        \midrule
        
        \multirow{5}{*}{\shortstack{Qwen2.5-\\Math-1.5B}} 
        & ORM & 50.73 & 54.80 & 56.80 & 58.07
        & 35.58 & 38.18 & 38.44 & \underline{40.17} \\
        & PRM & 51.80 & 55.73 & 56.87 & 58.00
        & 34.72 & 37.84 & 38.70 & 38.96 \\
        & PQM & \underline{52.67} & \underline{56.60} & \underline{58.87} & \underline{58.80}
        & \underline{36.88} & \underline{38.61} & \underline{40.61} & \underline{39.83} \\
        & IPRM & 44.27 & 47.27 & 48.33 & 47.47 & 32.55 & 34.46 & 35.32 & 34.55 \\
        \rowcolor{blue!10}\cellcolor{white} & \textbf{\method (ours)} & \textbf{54.07} & \textbf{58.40} & \textbf{61.00} & \textbf{63.00}
        & \textbf{38.70} & \textbf{39.74} & \textbf{41.04} & \textbf{43.55}\\
        
        \midrule
        
        \multirow{5}{*}{\shortstack{Qwen2.5-\\Math-7B}} 
        & ORM & 51.87 & 57.67 & 59.47 & 60.73 & 37.49 & 40.26 & \underline{44.07} & 43.72 \\
        & PRM & 52.13 & 55.67 & 59.93 & 60.13 & 37.58 & 40.52 & 41.04 & \underline{43.81} \\
        & PQM & \underline{52.73} & \underline{57.87} & \underline{59.20} & \underline{61.13} & \underline{37.84} & \underline{40.61} & 42.60 & 43.29 \\
        & IPRM & 49.53 & 54.07 & 54.20 & 52.60 & 35.67 & 38.53 & 40.26 & 39.57 \\
        \rowcolor{blue!10}\cellcolor{white} & \textbf{\method (ours)} & \textbf{56.07} & \textbf{60.60} & \textbf{62.87} & \textbf{64.07} & \textbf{39.83} & \textbf{42.77} & \textbf{46.49} & \textbf{48.40} \\

        \midrule
        
        \multirow{5}{*}{\shortstack{Llama3.1-\\8B}} 
        & ORM & 38.13 & 38.80 & 38.93 & 36.67 & 25.63 & 26.93 & \textbf{29.35} & 27.62 \\
        
        & PRM & 37.87 & 39.67 & 40.13 & 39.53 & \underline{26.84} & \textbf{28.66} & 27.97 & 27.97 \\
        & PQM & \underline{39.20} & \underline{40.47} & \underline{41.00} & \textbf{41.27} & 26.58 & 27.71 & 28.31 & \underline{28.23} \\
        & IPRM & 37.07 & 38.67 & 37.13 & 34.13 & 26.49 & 25.89 & 26.15 & 24.42 \\
        \rowcolor{blue!10}\cellcolor{white} & \textbf{\method (ours)} & \textbf{40.20} & \textbf{41.00} & \textbf{42.07} & \underline{41.00} & \textbf{26.93} & \underline{28.40} & \underline{28.74} & \textbf{29.96} \\
        \bottomrule
        \end{tabular}
    }
    \label{Beam Search}
    \vspace{-10pt}
\end{table}

\subsection{Beam search Experiments}
\label{sec:Beam Search}
We conduct beam search on MATH500 and an out-of-domain (OOD) dataset Gaokao2023 \citep{liao2024mario} to evaluate the capacity of reward models in guiding LLM reasoning via step-level rewards. For each question, beam search initiates by sampling $N$ responses. Subsequently, a beam of $b$ candidates with the highest rewards is maintained and expanded during the generation process. Our CRM computes $S(t)$ as the step-level reward.
We report the best accuracy achieved across various total sampling sizes $N$, with the results averaged over three random seeds. Detailed experimental settings are provided in Appendix \ref{appendix:implementation_details}.

\paragraph{CRM provides effective and consistent step-level guidance for beam search.}
As shown in Table \ref{Beam Search}, our \method achieves the highest accuracy on both datasets when using the Qwen2.5-Math-1.5B and Qwen2.5-Math-7B and demonstrates the best performance in the majority of cases for the Llama3.1-8B. Notably, the performance of CRM scales effectively with the total sampling size $N$. As $N$ increases from $4$ to $100$, the performance gap over baseline methods widens, underscoring the scalable advantage of \method in selecting more promising intermediate steps within larger search spaces.
This can be attributed to how the reward model guides the beam search algorithm. Specifically, beam search relies on the reward model to prune a vast number of trajectories by performing two distinct types of comparison: (i) ranking trajectories that share a common prefix, and (ii) ranking entirely distinct reasoning paths (cross-sample). By framing the reward as the conditional probability of reaching the correct answer given current partial trajectory, our consistent probabilistic modeling provides meaningful step-level rewards applicable to both ranking scenarios, thereby effectively guiding the reasoning process.

\subsection{RL Optimization Experiments}
\label{sec:RL Optimization}
\begin{table}[ht]
\caption{Pass@1 accuracy evaluated on six mathematical reasoning benchmarks.}
\label{tab:RL}
\centering
\renewcommand{\arraystretch}{1.2}
\resizebox{\linewidth}{!}{
\begin{tabular}{llcccccc}
\toprule
\textbf{\makecell{VR from outcome \\ ground-truth}} & \textbf{Method} & \textbf{\makecell{MATH\\500}} & \textbf{\makecell{Minerva\\Math}} & \textbf{\makecell{Olympiad\\Bench}} & \textbf{AIME25} & \textbf{AIME24} & \textbf{AMC23} \\
\midrule
\multirow{4}{*}{\makecell{VR Disabled}} 
  & PURE                     & 76.0 & 30.8 & 36.7 & 13.3 & 26.6 & \textbf{70.0} \\
  & PRM                      & 71.6 & 36.3 & 32.5 & 13.3 & 10.0 & 57.5 \\
  & PQM                      & 72.0 & 34.1 & 34.3 & 13.3 & 13.3 & 52.5 \\
  \rowcolor{blue!10}\cellcolor{white} & \textbf{\method (ours)}           & \textbf{77.8} & \textbf{40.0} & \textbf{39.3} & \textbf{23.3} & \textbf{43.3} & 67.5 \\
\midrule
\multirow{3}{*}{\makecell{VR Enabled}} 
  & Prime                    & 81.2 & 29.4 & 40.8 & 16.6 & 26.6 & \textbf{72.5} \\
  & PURE                     & \textbf{82.4} & 40.0 & 41.3 & 23.3 & 23.3 & 70.0 \\
  % & VR                     & 76.2 & 38.6 & 38.0 & 16.6 & 30.0 & 62.5 \\
  & VR                     & 76.2 & 38.6 & 38.0 & 16.6 & 30.0 & 62.5 \\
  \rowcolor{blue!10}\cellcolor{white} & \textbf{\method + VR}             & 80.4 & \textbf{43.0} & \textbf{42.1} & \textbf{26.6} & \textbf{33.3} & \textbf{72.5} \\
\bottomrule
\end{tabular}
}
\end{table}
During RL optimization, the reward model provides step-level dense rewards to guide the policy model toward generating improved reasoning trajectories.
This setting allows us to empirically validate the effectiveness and accuracy of credit assignment delivered by the reward model in practice.

Our \method provides step-wise process rewards $r_t = \log(1 - h(t))$. This reward formulation is not only theoretically justified in Section~\ref{sec:Modeling} but also empirically validated in Appendix~\ref{appendix:process_reward_formulation}.
We adopt RLOO (Leave-One-Out) \citep{ahmadian2024back} to estimate the advantage based on $r_t$.
Since RLOO is originally defined at the sequence level, following prior work \citep{cheng2025stop}, we employ its token-level variant.
We use Orz-Math-57k \citep{hu2025open} as the RL training set. 
All policy models are initialized from Qwen2.5-Math-7B and trained with full parameter optimization. 
Details of advantage estimation, policy updates, and implementation settings are provided in Appendix~\ref{appendix:implementation_details}.

We compare against PRM and PQM, and additionally include two stronger RL baselines that use dense rewards: Prime \citep{cui2025process} and PURE \citep{cheng2025stop}.
Prime adopts online reward model updates, jointly optimizing the reward model and policy, but its reliance on verifiers prevents adaptation to ground-truth–free settings.
PURE adopts min-form credit assignment by defining the value function as the minimum of future rewards.
However, as stated in its original paper, it becomes prone to reward hacking without verifiable rewards.
In addition, we also add the VR baseline, where the policy is trained using only final-answer verifiable rewards.
To comprehensively evaluate performance, we use six widely adopted benchmarks: AIME25 \citep{maa2025aime}, AIME24 \citep{maa2024aime}, AMC23 \citep{maa2023aime}, MATH500 \citep{hendrycks2021measuring}, Minerva Math \citep{lewkowycz2022solving}, and OlympiadBench \citep{he2024olympiadbench}.
We report the pass@1 accuracy under the zero-shot setting.

\paragraph{\method boosts RL performance without VR.}
\method-based RL functions without verifiable rewards \textbf{(VR Disabled)}. 
For fair comparison with VR-dependent methods, we additionally report results where rewards from \method are combined with verifiable rewards \textbf{(VR Enabled)}.
The results are shown in Table~\ref{tab:RL}.
In the VR Disabled setting, \method attains the best Pass@1 accuracy on the majority of benchmarks, significantly outperforming the baselines.
For instance, on AIME24 it reaches 43.3\%, exceeding PURE by +16.7.
CRM consistently outperforms the VR baseline.
When augmented with VR, \method + VR yields further gains and achieves the highest performance on most benchmarks. 
This suggests that the process rewards provided by \method are complementary to ground-truth–based verifiable rewards, rather than redundant.
Remarkably, even without VR, \method delivers performance comparable to VR-enabled methods.
The strength of \method lies in its explicit linkage between process rewards and outcomes, which ensures that the contribution of each step is causally aligned with the final result. 
Since process rewards quantify how much each step advances or undermines the probability of reaching the correct final answer, rather than relying on local heuristics, this design enables precise credit assignment. 
As a result, the dense and reliable reward signal facilitates effective RL optimization and remains robust to reward hacking, as further analyzed in Section~\ref{sec:reward_hacking_analysis}.

\subsection{Extended Analysis}
In this section, we conduct a more in-depth analysis of \method by investigating the following research questions.
\textbf{RQ1}: What does reward hacking manifest as, and how can it be mitigated?
\textbf{RQ2}: Does \method exhibit self-reflection behavior?
\textbf{RQ3}: How efficient is \method in utilizing supervision data?
\textbf{RQ4}: Can \method be applied to domains beyond mathematics?

\subsubsection{Reward Hacking in RL Optimization}
\label{sec:reward_hacking_analysis}
In RL optimization, employing a reward model is prone to reward hacking \citep{gao2024designing, cheng2025stop}.
We observe that reward hacking typically manifests as the generation of excessively repetitive content.
To capture this behavior, we introduce the repeat score, an n-gram–based \citep{li2015diversity} metric ranging from 0 to 1, where higher values indicate a greater degree of textual repetition.
The formal definition and examples of reward hacking are provided in Appendix \ref{appendix:reward_hacking}.

\begin{figure*}[t]
    \centering
    \includegraphics[width=\textwidth]{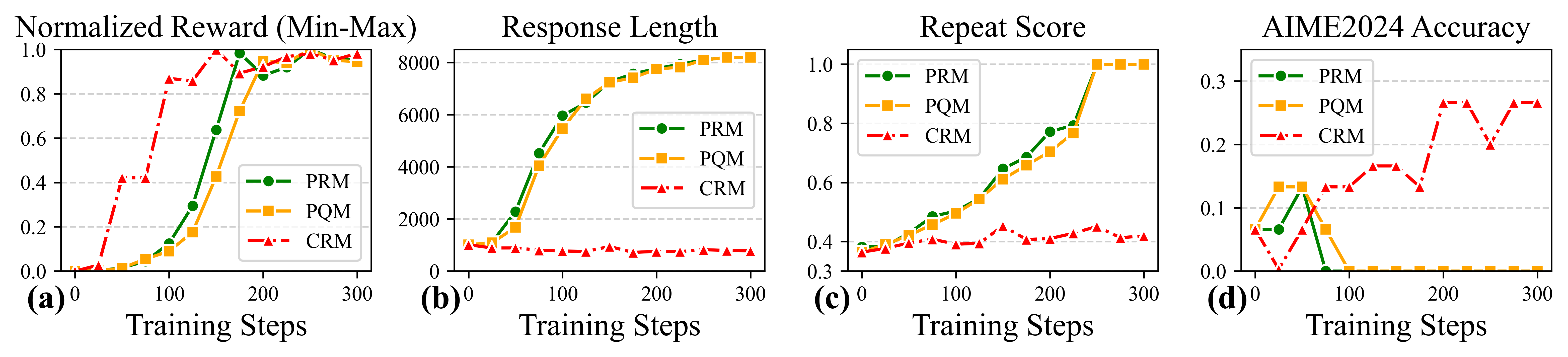}
    \caption{Evolution of (a) Normalized Reward (Min-Max), (b) Response Length, (c) Repeat Score, and (d) downstream task performance with training steps.}
    \label{fig:reward_hacking}
    \vspace{-10pt}
\end{figure*}

\textbf{Finding 1: Reward hacking manifests as a rapid reward increase accompanied by repetitive outputs and declining downstream performance (RQ1), whereas \method remains robust.}
Figure~\ref{fig:reward_hacking} reveals that under PRM and PQM, the reward quickly escalates despite declining downstream accuracy. 
This discrepancy arises from a pathological incentive, where models inflate rewards through excessively long, repetitive outputs, as reflected in their near-saturated repeat scores.
This indicates that the reward fails to discriminate between genuine reasoning step and superficial repetition.
In contrast, \method tightly couples intermediate rewards to the final outcome, ensuring that reward faithfully reflects reasoning quality (\textbf{RQ1}). 
\method's precise credit assignment stabilizes optimization and suppresses degenerate strategies, thereby enhancing robustness to reward hacking.

\subsubsection{Self-Reflection during Reasoning Process}
\begin{wrapfigure}[10]{r}{0.5\textwidth}
    \vspace{-0.5cm} 
    \centering
    \includegraphics[width=0.48\textwidth]{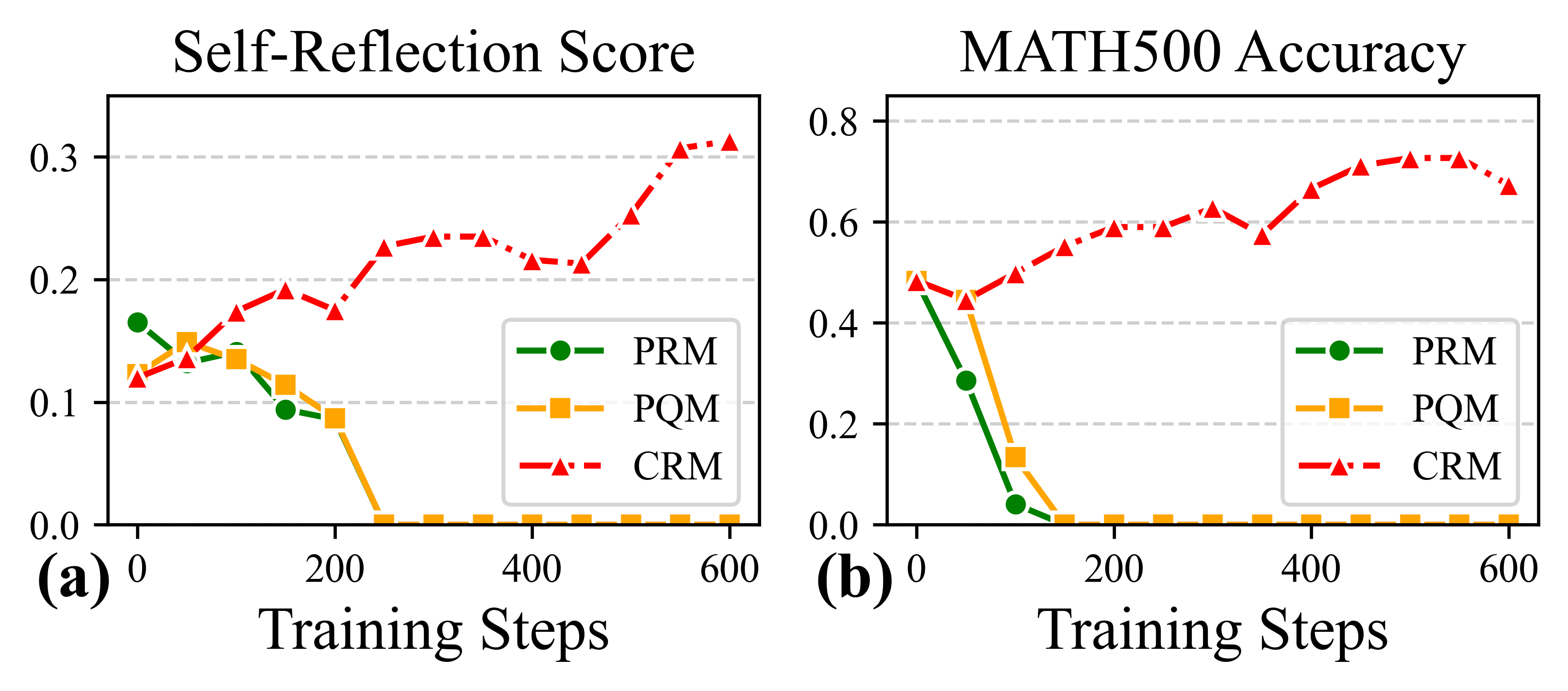}
    \vspace{-0.3cm}
    \caption{Evolution of self-reflection and downstream accuracy during RL training}
    \label{fig:self-reflection}
\end{wrapfigure}
Previous works \citep{he2024olympiadbench, liu2025understanding, bensal2025reflect} have observed that improvements in reasoning ability are often associated with a key phenomenon of self-reflection, in which models actively review and check previous steps.
This phenomenon provides important insights into understanding and enhancing reasoning in LLMs. 
Following prior research \citep{yeo2025demystifying, liu2025trust, liu2025understanding}, we define the self-reflection score as the average frequency of reflective expressions (e.g., “rethink”, “let’s check”) appearing in a model’s response, normalized by the output length (per 1000 tokens), which measures self-reflection capability during reasoning.
The full list of reflective expressions is provided in the Appendix \ref{appendix:reflective_expressions}.

\textbf{Finding 2: \method encourages more self-reflection behaviors \textbf{(RQ2)}.}
As shown in Figure~\ref{fig:self-reflection}, the self-reflection score of \method steadily rises during RL training, accompanied by improvements in downstream MATH500 accuracy, indicating that the model becomes increasingly reflective. 
In contrast, PRM and PQM show little to no growth in self-reflection and their MATH500 accuracy collapses early.
The temporal co-movement between rising self-reflection and improving MATH500 accuracy suggests that \method’s precise credit assignment fosters more meaningful reasoning behaviors, even without verifiable rewards.

\subsubsection{Ablation Study}
\begin{wraptable}[10]{r}{0.5\textwidth}
\centering
\caption{Ablation results for $\mathcal{L}_z$.}
\resizebox{0.48\textwidth}{!}{
\begin{tabular}{c|ccccc}
\hline
\makecell{Proportion of \\ data used for $\mathcal{L}_z$}   & @8   & @16  & @32  & @64  & @128  \\ \hline
0\%  & 47.0 & 44.2 & 41.6 & 39.2 & 38.2 \\
10\% & 52.4 & 51.2 & 50.6 & 49.6 & 47.6 \\
25\% & 54.0 & 52.4 & 53.6 & 53.2 & 52.0 \\
50\% & 54.4 & 53.6 & 57.2 & 55.8 & 55.0 \\
100\%& 53.0 & 56.4 & 56.6 & 55.8 & 56.6 \\ \hline
\end{tabular}
}
\label{tab:ablation_Lz}
\end{wraptable}
% \vspace{-10pt}
\textbf{Ablation on loss for \method training.} 
Our modeling approach employs three losses (Eq.~\ref{eq:L_S}, Eq.~\ref{eq:L_W}, Eq.~\ref{eq:L_z}). 
Among them, $\mathcal{L}_S$ and $\mathcal{L}_W$ are supervised by the label $l \in \{0, 1\}$ indicating whether the final answer is correct, with a one-to-one correspondence to samples. 
Removing these losses would reduce the number of training samples and undermine fair comparison. 
Therefore, we preserve $\mathcal{L}_S$ and $\mathcal{L}_W$ while conducting ablations on $\mathcal{L}_z$.
The loss $\mathcal{L}_z$ directly encourages the model to identify the exact step at which a wrong state occurs.
Specifically, we set the total amount of data in the dataset applicable to $\mathcal{L}_z$ as 100\% and apply scaling, evaluating \method with 0\%, 10\%, 25\%, and 50\% of the data under the Best-of-N sampling on the MATH500 dataset using Qwen2.5-3B-Instruct.

\textbf{Finding 3: \method exhibits high data efficiency \textbf{(RQ3)}.}
The ablation results in Table~\ref{tab:ablation_Lz} show that even a very small proportion of data applicable to $\mathcal{L}_z$ leads to substantial performance gains: moving from 0\% to just 10\% supervision produces a large improvement in Best-of-N accuracy. 
Increasing the proportion of data yields additional benefits, but the improvements quickly saturate, with 50\% already achieving near-optimal performance. 
This demonstrates that \method requires only limited data for $\mathcal{L}_z$ to achieve strong results, highlighting its high data utilization efficiency.
A more detailed analysis of the reasons behind this efficiency, as well as comparisons with the baseline, can be found in the Appendix ~\ref{appendix:data_efficiency}.

\subsubsection{Validation on More Domains}
To further validate the effectiveness of our \method, we extend our evaluation to additional domains beyond mathematics. 
We leverage the multi-domain dataset MMLU-Pro-CoT-Train from VersaPRM \citep{zeng2025versaprm} as it provides reasoning processes with step-wise labels across a wide range of domains. 
We train the reward models including \method and the baselines on this dataset using Qwen2.5-3B-Instruct as backbone and subsequently assess their performance using Best-of-N sampling on five distinct domains including biology, business, health, history, and physics.

\textbf{Finding 4: \method can be extended to other domains and demonstrates strong performance. \textbf{(RQ4)}.}
As shown in Table \ref{tab:domain}, \method outperforms the baselines in almost all tested domains. 
This demonstrates that \method's validity is not limited to mathematical tasks and can be applied to a broader range of domains.
\begin{table}[h]
    \centering
    \caption{Best-of-N accuracy comparison across different domains (MMLU-Pro-CoT-Eval).}
    \label{tab:domain}
    \resizebox{\textwidth}{!}{
    \begin{tabular}{l|ccc|ccc|ccc|ccc|ccc}
        \toprule
        \multirow{2}*{Methods} & \multicolumn{3}{c|}{Biology} & \multicolumn{3}{c|}{Business} & \multicolumn{3}{c|}{Health} & \multicolumn{3}{c|}{History} & \multicolumn{3}{c}{Physics} \\
        & @16 & @32 & @64 & @16 & @32 & @64 & @16 & @32 & @64 & @16 & @32 & @64 & @16 & @32 & @64 \\
        \midrule
        PRM & 80.0 & 79.2 & 79.2 & 64.8 & 64.8 & 66.2 & 60.7 & 57.1 & 57.1 & 48.0 & 48.7 & 48.7 & 61.0 & 58.9 & 60.3 \\
        PQM & 80.8 & 80.0 & 80.8 & 64.1 & 63.5 & 65.5 & 62.1 & 61.4 & 61.4 & 50.0 & 51.3 & 49.3 & 61.6 & 58.2 & 62.3 \\
        \rowcolor{blue!10}\textbf{\method (ours)} & 83.1 & 83.1 & 82.3 & 66.2 & 66.9 & 68.2 & 61.4 & 62.1 & 60.7 & 50.7 & 52.0 & 51.3 & 64.4 & 63.7 & 66.4 \\
        \bottomrule
    \end{tabular}
    }
\end{table}

\section{Conclusion}
In this paper, we introduce \method, which frames LLM reasoning as a temporal probabilistic process.
By explicitly modeling the causal dependencies between steps and linking process to the outcome, \method addresses the limitations of isolated step modeling and limited outcome awareness in prior work.
Experiments across Best-of-N sampling, beam search, and RL demonstrate that \method consistently outperforms strong baselines, improves cross-sample comparability, and is robust to reward hacking.
Building on these results, we plan to extend \method to broader domains and task formats, and we hope this work catalyzes further research on reward-model–driven RL for reasoning, moving beyond reliance on ground truth to achieve broader generalization.

\section{Acknowledgments}
The research received partial support from National Natural Science Foundation of China (Grant No. 62406193).
The authors also gratefully acknowledge further assistance provided by Shanghai Frontiers Science Center of Human-centered Artificial Intelligence, MoE Key Lab of Intelligent Perception and Human-Machine Collaboration, and HPC Platform of ShanghaiTech University.

\bibliography{iclr2026_conference}
\bibliographystyle{iclr2026_conference}

\newpage
\appendix
\section{Implementation Details}
\label{appendix:implementation_details}

All experiments are conducted on 8 NVIDIA H20 GPUs.
\subsection{Reward Model Training}
For data processing, we follow prior work \citep{yu2023ovm, zheng2024processbench, zhang2025lessons}, using newline characters to indicate boundaries between individual reasoning steps. 
In terms of implementation, we used the ZeRO-2 optimization stage of DeepSpeed with bfloat16 precision to train the model. We employed AdamW optimizer with a learning rate of 5e-6 and a batch size of 32.

\subsection{Best-of-N sampling}
For the Best-of-N sampling evaluation, we adopt Llama-3.1-8B-Instruct as the generator and generate trajectories using the vLLM pipeline with temperature=1, top-p=1, and a maximum length of 2048. Each generated trajectory is subsequently scored by the reward models. Consistent with the settings in their original papers and code implementations, PRM, PQM, and IPRM aggregate process rewards into a sequence-level score by taking the minimum value across all steps. In contrast, our \method computes $S(T)$ as the trajectory-level score.

\subsection{Beam Search}
Our beam search is configured with four total sampling numbers, denoted as $N\in\{4, 8, 20, 100\}$. For each value of $N$, we test three distinct beam sizes, $b$. Initially, $N$ candidate responses are generated from a given question. In each subsequent step, the reward model evaluates the current set of candidates and selects the top-scoring $b$ trajectories as prefixes for expansion, where each prefix produces $N/b$ new continuations. This iterative process continues until all trajectories have generated a complete answer or the predefined maximum step count or token length is reached. The trajectory with the highest reward is then selected to extract the final answer. 

For the evaluation of the Qwen2.5-Math-1.5B and Qwen2.5-Math-7B reward models, we adopt Qwen2.5-3B as the generator, and we adopt Llama3.1-8B as the generator for its corresponding reward model. All rollout generation is performed using the vLLM pipeline, with a temperature of 0.7 and top-p of 1. We set a maximum token length of 4096 and a maximum of 30 steps for each trajectory generation.

\subsection{RL Optimization}
\paragraph{Advantage Estimation.}
The reward model provides step-wise process rewards $r$, and we adopt RLOO (Leave-One-Out) \citep{ahmadian2024back} to estimate the advantage based on these rewards.
RLOO uses multiple model outputs other than the current one to compute a baseline, effectively reducing variance in advantage estimation.
Since RLOO is originally defined at the sequence level, following prior work \citep{cheng2025stop}, we employ its token-level variant.
Specifically, given a question, the LLM generates $K$ responses $\{ y_i \}_{i=1}^K$, each containing at most $M$ tokens.
Let $r_i^t$ denote the reward of the $t$-th token in the $i$-th response, and $A_i^t$ denote its corresponding advantage.
In our implementation, we set the discount factor $\gamma$ to 1.
The advantage is computed as:
\begin{equation}
A_i^t = 
\sum_{j=t}^{M} r_i^j
\; - \;
\frac{ \sum_{k \ne i} \sum_{l=1}^{M} \sum_{j=l}^{M} r_k^j}
{(K - 1)M}
\end{equation}

\paragraph{Policy Update.}
We optimize the policy model $\pi_{\theta}$ by maximizing the following objective:
\begin{equation}
    L(\theta) = \mathbb{E}_{i,t} \\
    \left[
        \min \left(
            \frac{\pi_\theta(y_i^t \mid y_i^{<t})}{\pi_{\theta_{\text{old}}}(y_i^t \mid y_i^{<t})} A_i^t,\,
            \text{clip}\!\bigg(
                \frac{\pi_\theta(y_i^t \mid y_i^{<t})}{\pi_{\theta_{\text{old}}}(y_i^t \mid y_i^{<t})},
                1 - \epsilon,\, 1 + \epsilon
            \bigg) A_i^t
        \right)
        - \beta \mathbb{D}_\text{KL} (\pi_{\theta} || \pi_{ref} )
    \right]
\end{equation}
where $\pi_{\theta_{\text{old}}}$ is the old policy and $\pi_{\theta_{\text{ref}}}$ is the reference model.
$\epsilon$ is the PPO clipping threshold and $\beta$ controls the strength of the KL penalty; both are hyperparameters.

\paragraph{Experimental setting.}
We use Orz-Math-57k \citep{hu2025open} as the RL training set. 
The dataset was created through a thorough data-cleaning process, ensuring no overlap with commonly used benchmarks.
Experiments are conducted with veRL \citep{sheng2024hybridflow} framework.
We segment each LLM-generated response into steps using double line breaks and assign scores to each step with the reward model.
We train with a fixed learning rate of 1e-6, a prompt batch size of 64, and 4 responses sampled per prompt, using a KL coefficient $\beta$ = 1e-3 and a clip ratio $\epsilon$ = 0.2. 
For generation, we adopt vLLM with temperature = 1, top-p=1, and a maximum length of 8192 tokens, while during testing we set temperature =0 and top-p=1.

\paragraph{CRM+VR setting.}
When used CRM alone, it provides step-wise rewards ($r_t^{C}$) for all steps, including the final reasoning step $T$. When combined with VR, only the final step additionally receives a verifiable reward ($r_T^{V}$), while earlier steps still rely solely on CRM. 
Formally, the reward at each step $t$ is defined as:
\begin{equation}
r_t =
\begin{cases}
r_t^{C}, & \text{if } t < T \\
r_T^{C} + r_T^{V}, & \text{if } t = T
\end{cases}
\end{equation}

\section{Full Experimental Results for Beam Search}
The detailed specification of the beam sizes $b$ used for each $N$, along with the complete results, is provided in Table \ref{bs_1.5B}, Table \ref{bs_7B} and Table \ref{bs_8B}.

\begin{table}[ht]
\centering
\caption{Beam Search performance of Qwen2.5-Math-1.5B on MATH500 and Gaokao2023}
\label{bs_1.5B}
\resizebox{\textwidth}{!}{
    \begin{tabular}{cc|ccccc|ccccc}
        \toprule
        \multirow{2}*{$N$} & \multirow{2}*{$b$} & \multicolumn{5}{c|}{\textbf{MATH500}} & \multicolumn{5}{c}{\textbf{GAOKAO2023}} \\
        &
        & ORM & PRM & PQM & IPRM & CRM
        & ORM & PRM & PQM & IPRM & CRM \\
        
        \midrule
        
        \multirow{3}*{4} & 4 & 49.20 $\pm$ 0.60 & 49.93 $\pm$ 1.07 & 50.60 $\pm$ 0.80 & 44.27 $\pm$ 0.73 & 50.40 $\pm$ 0.20 & 34.46 $\pm$ 0.35 & 34.29 $\pm$ 1.30 & 35.06 $\pm$ 1.30 & 32.55 $\pm$ 1.47 & 34.03 $\pm$ 2.34\\
        & 2 & 50.73 $\pm$ 1.47 & 51.80 $\pm$ 0.80 & 52.60 $\pm$ 0.60 & 44.13 $\pm$ 1.27 & 53.27 $\pm$ 1.93 & 35.58 $\pm$ 1.82 & 34.55 $\pm$ 1.82 & 34.55 $\pm$ 1.56 & 31.00 $\pm$ 0.43 & 38.70 $\pm$ 0.26 \\
        & 1 & 49.73 $\pm$ 2.27 & 50.27 $\pm$ 3.33 & 52.67 $\pm$ 0.53 & 43.67 $\pm$ 2.33 & 54.07 $\pm$ 2.73 & 32.99 $\pm$ 2.08 & 34.72 $\pm$ 1.39 & 36.88 $\pm$ 0.78 & 29.96 $\pm$ 1.21 & 36.71 $\pm$ 0.95 \\

        \midrule 
        
        \multirow{3}*{8} & 8 & 51.27 $\pm$ 0.33 & 52.53 $\pm$ 1.07 & 54.60 $\pm$ 2.20 & 47.27 $\pm$ 0.93 & 53.67 $\pm$ 1.53 & 38.18 $\pm$ 1.30 & 35.15 $\pm$ 1.73 & 37.58 $\pm$ 1.65 & 34.46 $\pm$ 0.87 & 36.54 $\pm$ 2.42 \\
        & 4 & 54.80 $\pm$ 1.40 & 55.73 $\pm$ 1.67 & 56.20 $\pm$ 0.60 & 46.07 $\pm$ 1.33 & 57.47 $\pm$ 0.13 & 37.49 $\pm$ 1.47 & 37.84 $\pm$ 1.39 & 37.06 $\pm$ 0.35 & 32.38 $\pm$ 0.35 & 39.31 $\pm$ 1.47 \\
        & 2 & 52.80 $\pm$ 0.60 & 54.13 $\pm$ 0.47 & 56.60 $\pm$ 0.80 & 45.27 $\pm$ 1.53 & 58.40 $\pm$ 1.20 & 36.28 $\pm$ 0.35 & 36.54 $\pm$ 1.90 & 38.61 $\pm$ 0.61 & 30.56 $\pm$ 0.35 & 39.74 $\pm$ 0.52 \\

        \midrule
        
        \multirow{3}*{20} & 20 & 53.93 $\pm$ 0.67 & 55.73 $\pm$ 0.67 & 57.00 $\pm$ 0.20 & 47.73 $\pm$ 1.47 & 55.40 $\pm$ 1.40 & 37.75 $\pm$ 0.43 & 37.75 $\pm$ 0.43 & 39.05 $\pm$ 0.69 & 34.29 $\pm$ 1.30 & 36.62 $\pm$ 2.34 \\
        & 10 & 56.80 $\pm$ 0.60 & 56.87 $\pm$ 0.53 & 58.87 $\pm$ 0.73 & 48.33 $\pm$ 1.87 & 59.80 $\pm$ 0.20 & 38.44 $\pm$ 1.04 & 38.53 $\pm$ 1.21 & 39.74 $\pm$ 2.08 & 35.32 $\pm$ 0.52 & 41.04 $\pm$ 0.78 \\
        & 5 & 55.00 $\pm$ 0.40 & 55.80 $\pm$ 1.40 & 57.47 $\pm$ 1.53 & 47.07 $\pm$ 0.53 & 61.00 $\pm$ 0.60 & 37.66 $\pm$ 0.52 & 38.70 $\pm$ 0.52 & 40.61 $\pm$ 1.73 & 31.69 $\pm$ 0.78 & 40.78 $\pm$ 0.26 \\

        \midrule
        
        \multirow{3}*{100} & 50 & 57.00 $\pm$ 1.00 & 57.80 $\pm$ 0.60 & 58.80 $\pm$ 0.60 & 46.67 $\pm$ 1.33 & 63.00 $\pm$ 0.40 & 40.17 $\pm$ 0.87 & 38.53 $\pm$ 0.95 & 39.39 $\pm$ 1.13 & 34.55 $\pm$ 1.30 & 39.91 $\pm$ 1.13 \\
        & 25 & 58.07 $\pm$ 0.53 & 57.40 $\pm$ 0.40 & 57.73 $\pm$ 0.27 & 47.47 $\pm$ 1.13 & 61.47 $\pm$ 0.93 & 38.87 $\pm$ 0.87 & 38.96 $\pm$ 1.30 & 39.83 $\pm$ 1.73 & 32.99 $\pm$ 1.56 & 43.55 $\pm$ 0.61 \\
        & 10 & 55.87 $\pm$ 0.33 & 58.00 $\pm$ 1.20 & 57.67 $\pm$ 1.53 & 46.13 $\pm$ 0.27 & 61.87 $\pm$ 0.73 & 38.70 $\pm$ 0.26 & 38.27 $\pm$ 0.95 & 39.05 $\pm$ 0.43 & 32.99 $\pm$ 0.26 & 42.08 $\pm$ 1.04 \\
         
        \bottomrule
    \end{tabular}
}
\end{table}

\begin{table}[ht]
\centering
\caption{Beam Search performance of Qwen2.5-Math-7B on MATH500 and Gaokao2023}
\label{bs_7B}
\resizebox{\textwidth}{!}{
    \begin{tabular}{cc|ccccc|ccccc}
        \toprule
        \multirow{2}*{$N$} & \multirow{2}*{$b$} & \multicolumn{5}{c|}{\textbf{MATH500}} & \multicolumn{5}{c}{\textbf{GAOKAO2023}} \\
        &
        & ORM & PRM & PQM & IPRM & CRM
        & ORM & PRM & PQM & IPRM & CRM \\
        
        \midrule
        
        \multirow{3}*{4} & 4 & 51.47 $\pm$ 1.53 & 51.20 $\pm$ 1.40 & 50.87 $\pm$ 0.73 & 49.53 $\pm$ 1.07 & 51.20 $\pm$ 1.00 & 36.88 $\pm$ 0.52 & 34.55 $\pm$ 0.52 & 36.45 $\pm$ 1.73 & 35.67 $\pm$ 1.47 & 34.72 $\pm$ 0.35 \\
        & 2 & 51.87 $\pm$ 0.73 & 52.13 $\pm$ 1.47 & 52.73 $\pm$ 0.67 & 45.73 $\pm$ 1.67 & 54.20 $\pm$ 0.60 & 37.49 $\pm$ 2.25 & 35.32 $\pm$ 0.26 & 37.84 $\pm$ 0.61 & 31.95 $\pm$ 1.56 & 38.53 $\pm$ 1.47 \\
        & 1 & 51.53 $\pm$ 1.87 & 50.93 $\pm$ 0.47 & 52.27 $\pm$ 1.53 & 42.13 $\pm$ 0.87 & 56.07 $\pm$ 0.53 & 35.76 $\pm$ 1.13 & 37.58 $\pm$ 1.90 & 37.40 $\pm$ 0.52 & 32.29 $\pm$ 0.95 & 39.83 $\pm$ 1.73 \\

        \midrule 
        
        \multirow{3}*{8} & 8 & 54.60 $\pm$ 0.40 & 53.60 $\pm$ 0.60 & 56.53 $\pm$ 1.67 & 54.07 $\pm$ 0.13 & 56.00 $\pm$ 0.40 & 38.27 $\pm$ 1.21 & 37.32 $\pm$ 1.39 & 37.58 $\pm$ 0.87 & 38.53 $\pm$ 1.99 & 38.87 $\pm$ 0.35 \\
        & 4 & 57.67 $\pm$ 1.33 & 55.67 $\pm$ 0.13 & 57.87 $\pm$ 1.53 &  50.27 $\pm$ 0.33 & 60.60 $\pm$ 1.40 & 40.26 $\pm$ 0.26 & 40.52 $\pm$ 0.78 & 39.65 $\pm$ 1.39 & 34.03 $\pm$ 1.04 & 42.60 $\pm$ 1.04 \\
        & 2 & 56.27 $\pm$ 0.93 & 54.07 $\pm$ 1.13 & 56.07 $\pm$ 0.53 & 46.33 $\pm$ 1.07 & 58.27 $\pm$ 2.33 & 39.57 $\pm$ 1.47 & 39.13 $\pm$ 1.90 & 40.61 $\pm$ 0.69 & 31.69 $\pm$ 0.52 & 42.77 $\pm$ 1.13 \\

        \midrule
        
        \multirow{3}*{20} & 20 & 56.27 $\pm$ 1.13 & 56.73 $\pm$ 0.27 & 56.93 $\pm$ 1.47 & 54.20 $\pm$ 2.40 & 57.80 $\pm$ 1.40 & 41.30 $\pm$ 1.30 & 41.04 $\pm$ 0.52 & 39.83 $\pm$ 0.69 & 40.26 $\pm$ 1.04 & 41.90 $\pm$ 0.69 \\
        & 10 & 59.47 $\pm$ 1.53 & 59.93 $\pm$ 0.87 & 59.07 $\pm$ 0.93 & 52.73 $\pm$ 1.67 & 61.40 $\pm$ 1.20 & 44.07 $\pm$ 0.61 & 40.78 $\pm$ 0.52 & 41.21 $\pm$ 0.61 & 37.14 $\pm$ 1.56 & 45.37 $\pm$ 0.61 \\
        & 5 & 58.80 $\pm$ 0.80 & 57.07 $\pm$ 0.53 & 59.20 $\pm$ 1.20 &  49.47 $\pm$ 1.13 & 62.87 $\pm$ 0.53 & 42.16 $\pm$ 2.51 & 40.61 $\pm$ 2.25 & 42.60 $\pm$ 0.52 & 35.41 $\pm$ 1.21 & 46.49 $\pm$ 0.52 \\

        \midrule
        
        \multirow{3}*{100} & 50 & 59.67 $\pm$ 1.73 & 58.40 $\pm$ 0.80 & 59.07 $\pm$ 0.53 & 52.60 $\pm$ 0.80 & 63.80 $\pm$ 1.00 & 43.55 $\pm$ 0.61 & 43.12 $\pm$ 1.30 & 42.68 $\pm$ 0.69 & 39.57 $\pm$ 2.51 & 48.31 $\pm$ 2.60 \\
        & 25 & 60.73 $\pm$ 0.67 & 60.13 $\pm$ 0.67 & 61.13 $\pm$ 0.87 & 51.73 $\pm$ 0.47 & 63.60 $\pm$ 1.00 & 43.64 $\pm$ 2.60 & 43.81 $\pm$ 0.35 & 43.03 $\pm$ 1.13 & 37.49 $\pm$ 0.95 & 47.62 $\pm$ 1.21 \\
        & 10 & 59.47 $\pm$ 0.53 & 59.27 $\pm$ 1.13 & 60.20 $\pm$ 1.80 & 48.80 $\pm$ 1.20 & 64.07 $\pm$ 0.33 & 43.72 $\pm$ 1.73 & 41.90 $\pm$ 0.43 & 43.29 $\pm$ 0.87 & 34.89 $\pm$ 0.95 & 48.40 $\pm$ 1.21 \\
         
        \bottomrule
    \end{tabular}
}
\end{table}

\begin{table}[ht]
\centering
\caption{Beam Search performance of Llama3.1-8B on MATH500 and Gaokao2023}
\label{bs_8B}
\resizebox{\textwidth}{!}{
    \begin{tabular}{cc|ccccc|ccccc}
        \toprule
        \multirow{2}*{$N$} & \multirow{2}*{$b$} & \multicolumn{5}{c|}{\textbf{MATH500}} & \multicolumn{5}{c}{\textbf{GAOKAO2023}} \\
        &
        & ORM & PRM & PQM & IPRM & CRM
        & ORM & PRM & PQM & IPRM & CRM \\
        
        \midrule
        
        \multirow{3}*{4} & 4 & 37.60 $\pm$ 1.00 & 37.67 $\pm$ 0.33 & 38.27 $\pm$ 0.93 & 37.07 $\pm$ 0.13 & 40.20 $\pm$ 0.40 & 25.37 $\pm$ 0.61 & 26.84 $\pm$ 1.47 & 25.37 $\pm$ 0.35 & 26.49 $\pm$ 1.56 & 26.93 $\pm$ 2.16 \\
        & 2 & 38.13 $\pm$ 0.47 & 37.67 $\pm$ 1.53 & 39.20 $\pm$ 0.80 & 34.27 $\pm$ 1.33 & 39.60 $\pm$ 0.80 & 25.63 $\pm$ 1.13 & 26.41 $\pm$ 0.87 & 26.58 $\pm$ 0.43 & 23.72 $\pm$ 1.73 & 25.89 $\pm$ 1.90 \\
        & 1 & 36.67 $\pm$ 1.53 & 37.87 $\pm$ 0.93 & 38.20 $\pm$ 0.60 & 32.73 $\pm$ 0.87 & 37.47 $\pm$ 0.93 & 23.55 $\pm$ 0.87 & 26.23 $\pm$ 1.82 & 25.54 $\pm$ 1.47 & 22.34 $\pm$ 0.52 & 25.80 $\pm$ 0.69 \\

        \midrule 
        
        \multirow{3}*{8} & 8 & 38.80 $\pm$ 1.80 & 39.07 $\pm$ 0.73 & 39.47 $\pm$ 1.93 & 38.67 $\pm$ 1.73 & 38.67 $\pm$ 0.53 & 25.45 $\pm$ 1.30 & 26.93 $\pm$ 0.35 & 27.71 $\pm$ 0.61 & 25.89 $\pm$ 2.42 & 28.40 $\pm$ 0.43 \\
        & 4 & 38.73 $\pm$ 1.87 & 39.07 $\pm$ 0.33 & 40.47 $\pm$ 1.13 & 36.73 $\pm$ 1.27 & 41.00 $\pm$ 1.40 & 26.93 $\pm$ 0.87 & 28.23 $\pm$ 0.35 & 26.93 $\pm$ 3.20 & 23.98 $\pm$ 2.25 & 26.84 $\pm$ 0.43 \\
        & 2 & 38.07 $\pm$ 0.53 & 39.67 $\pm$ 1.33 & 40.27 $\pm$ 0.73 & 34.40 
        $\pm$ 0.60 & 39.20 $\pm$ 0.80 & 24.07 $\pm$ 2.94 & 28.66 $\pm$ 1.99 & 26.32 $\pm$ 0.69 & 23.46 $\pm$ 0.95 & 26.15 $\pm$ 0.87 \\

        \midrule
        
        \multirow{3}*{20} & 20 & 36.33 $\pm$ 1.07 & 40.13 $\pm$ 1.87 & 41.00 $\pm$ 2.80 & 37.13 $\pm$ 0.87 & 40.53 $\pm$ 0.67 & 26.75 $\pm$ 0.52 & 27.27 $\pm$ 1.04 & 28.31 $\pm$ 1.30 & 26.15 $\pm$ 2.16 & 27.62 $\pm$ 0.69 \\
        & 10 & 38.93 $\pm$ 1.07 & 39.80 $\pm$ 2.40 & 40.67 $\pm$ 0.73 & 34.07 $\pm$ 1.33 & 42.07 $\pm$ 1.13 & 29.35 $\pm$ 0.52 & 27.53 $\pm$ 1.82 & 26.75 $\pm$ 1.30 & 24.68 $\pm$ 2.08 & 28.31 $\pm$ 1.04 \\
        & 5 & 37.73 $\pm$ 1.47 & 39.80 $\pm$ 1.40 & 40.40 $\pm$ 1.20 & 33.93 $\pm$ 1.07 & 40.80 $\pm$ 0.40 & 26.23 $\pm$ 0.78 & 27.97 $\pm$ 1.65 & 27.36 $\pm$ 0.95 & 25.02 $\pm$ 4.33 & 28.74 $\pm$ 0.35 \\

        \midrule
        
        \multirow{3}*{100} & 50 & 36.67 $\pm$ 0.73 & 39.07 $\pm$ 1.53 & 41.27 $\pm$ 1.53 & 32.20 $\pm$ 1.20 & 40.67 $\pm$ 0.33 & 27.62 $\pm$ 1.73 & 27.79 $\pm$ 0.78 & 28.23 $\pm$ 0.09 & 23.46 $\pm$ 0.17 & 28.31 $\pm$ 2.08 \\
        & 25 & 36.67 $\pm$ 1.73 & 39.53 $\pm$ 0.67 & 40.27 $\pm$ 0.33 & 33.13 $\pm$ 1.07 & 40.20 $\pm$ 0.60 & 27.19 $\pm$ 1.39 & 27.97 $\pm$ 1.13 & 27.53 $\pm$ 0.78 & 24.42 $\pm$ 0.52 & 29.96 $\pm$ 0.69 \\
        & 10 & 35.33 $\pm$ 0.87 & 38.47 $\pm$ 0.33 & 38.47 $\pm$ 0.33 & 34.13 $\pm$ 1.67 & 41.00 $\pm$ 1.40 & 27.27 $\pm$ 1.82 & 27.97 $\pm$ 0.87 & 27.01 $\pm$ 1.04 & 22.94 $\pm$ 1.21
        & 28.83 $\pm$ 1.56 \\
         
        \bottomrule
    \end{tabular}
}
\end{table}

\section{Reward Hacking}
\label{appendix:reward_hacking}
\subsection{Repeat Score}
To quantify the degree of redundancy in generated responses, we define the \textit{repeat score} based on $n$-gram statistics. 
This metric is a widely used indicator of textual repetition in the literature. 
Given a text sequence, we first normalize it by lowercasing, unifying quotation marks, and removing code blocks, LaTeX environments, and inline formulas. 
The text is then tokenized into alphanumeric tokens and basic punctuation. 

For an $n$-gram size $n$, we compute the \textit{Distinct-$n$} ratio \citep{li2015diversity}:
\[
\text{Distinct-}n = \frac{|\{\text{unique } n\text{-grams}\}|}{\text{total } n\text{-grams}}.
\]
The \textit{Repetition-$n$} score is defined as:
\[
\text{Repetition-}n = 1 - \text{Distinct-}n,
\]
which ranges from 0 to 1, with larger values indicating a higher proportion of repeated $n$-grams. 

In practice, since $n=2$ and $n=3$ are widely adopted choices in NLP research and have been shown to be effective for capturing local redundancy patterns, we follow this design in our implementation and compute $\textit{Repetition-2}$ and $\textit{Repetition-3}$, taking their maximum as the final \textit{repeat score}:
\[
\text{Repeat Score} = \max(\text{Repetition-2}, \text{Repetition-3}).
\]

\subsection{Examples of Reward Hacking}

\begin{tcolorbox}[title=Example 1,
  colback=gray!5,
  colframe=black!40,
  coltitle=black,
  fonttitle=\bfseries,
  breakable]

\textbf{Question:}\\[0.25em]
Given that the sum of the first $n$ terms of an arithmetic sequence $\{a_n\}$ is $S_n$, if $a_4 + a_6 = 18$, then $S_9$ equals (\,\;).\\[0.25em]
A:\,$99$\quad B:\,$81$\quad C:\,$72$\quad D:\,$63$

\vspace{1em}

\textbf{LLM Response:}\\[0.25em]
To solve this problem, we need to use the properties of an arithmetic sequence. Let's break down the problem step-by-step.

1. Arithmetic Sequence Formula:
The \(n\)-th term of an arithmetic sequence can be written as:
\[
a_n = a_1 + (n-1)d
\]

2. Sum of the First \(n\) Terms:
\[
S_n = \frac{n}{2} (2a_1 + (n-1)d)
\]

3. Given Condition:
We are given that \(a_4 + a_6 = 18\). Using the formula:
\[
a_4 = a_1 + 3d, \quad a_6 = a_1 + 5d
\]
So,
\[
a_4 + a_6 = 2a_1 + 8d = 18 \implies a_1 + 4d = 9
\]

4. Sum of the First 9 Terms:
\[
S_9 = \frac{9}{2} (2a_1 + 8d) = \frac{9}{2} (2a_1 + 8d) = 9 \cdot 99
\]

Let's solve this problem step-by-step.

1. Arithmetic Sequence Formula:

2. Sum of the first 9 terms: 
\quad \textcolor{red}{\(\longleftarrow\) Start generating repeated content.}

3. Sum of the first 9 terms:

4. Sum of the first 9 terms:

...

Let's solve the first 9 terms:

Let's solve the first 9 terms:

Let's solve the first 9 terms:

...

Let

Let9

Let9

Let

Let

Let9

Let9

...
\end{tcolorbox}

\begin{tcolorbox}[title=Example 2,
  colback=gray!5,
  colframe=black!40,
  coltitle=black,
  fonttitle=\bfseries,
  breakable]

\textbf{Question:}\\[0.25em]
Among the $900$ residents of Aimeville, there are $195$ who own a diamond ring, $367$ who own a set of golf clubs, and $562$ who own a garden spade. In addition, each of the $900$ residents owns a bag of candy hearts. There are $437$ residents who own exactly two of these things, and $234$ residents who own exactly three of these things. Find the number of residents of Aimeville who own all four of these things.

\vspace{1em}

\textbf{LLM Response:}\\[0.25em]
To solve this problem, we can use the principle of inclusion and exclusion. Let's define the following sets:

 - \(D\) = the set of residents who own a diamond ring.
 
 - \(G\) = the set of residents who own a set of golf clubs.
 
 - \(S\) = the set of residents who own a garden spade.
 
 - \(C\) = the set of residents who own a bag of candy hearts (which is the entire population of Aimeville).

We are given the following information:

 - \(|D| = 195\)
 
 - \(|G| = 367\)
 
 - \(|S| = 562\)
 
 - \(|C| = 900\)
 
The number of residents who own exactly two of these things is 437.

The number of residents who own exactly three of these things is 234.

We need to find the number of residents who own all four things, i.e., \(|D \cap G \cap S \cap C|\).

First, let's denote the number of residents who own exactly \(k\) of these things by \(N_k\). We know that:

\[N_2 = 437\]
\[N_3 = 234\]

...

Find the values of f(x) in [-pi / 4]
\quad \textcolor{red}{\(\longleftarrow\) Start generating repeated content.}

Find the values of f(x) in [-pi / 4]

Find the values of f(x) in [-pi / 4]

...

Find the maximum and minimum values of f(x) in [-pi / 4]

Find the maximum and minimum values of f(x) in [-pi / 4]

Find the maximum and minimum values of f(x) in [-pi / 4]

Find the maximum and minimum values of f(x) in [-pi / 4]

...

The values of f(x) in

The values of f(x) in

The values of f(x) in

The values of f(x) in

...
\end{tcolorbox}

\section{Reflective Expressions}
\label{appendix:reflective_expressions}
Table~\ref{tab:reflective_expressions} lists the reflective expressions used to compute the self-reflection score.  

\begin{table}[h]
\centering
\caption{Reflective expressions used in the computation of the self-reflection score.}
\begin{tabular}{ll}
\toprule
\textbf{Reflective Expressions} & \textbf{Example} \\
\midrule
\texttt{wait} & "wait, let me think" \\
\texttt{recheck} & "recheck this step" \\
\texttt{retry} & "retry the calculation" \\
\texttt{try again} & "let's try again" \\
\texttt{alternatively} & "alternatively, we can ..." \\
\texttt{however} & "however, this may fail" \\
\texttt{rethink} & "rethink the argument" \\
\texttt{let's check} & "let's check the result" \\
\texttt{let's verify} & "let's verify the answer" \\
\bottomrule
\end{tabular}
\label{tab:reflective_expressions}
\end{table}

\section{Cross-sample Comparison Capability}
\label{appendix:cross-sample}

\paragraph{PQM focuses on intra-sample ranking but lacks cross-sample comparability.} 
The core modeling principle of PQM \citep{li2024process} focuses on the relative ordering of steps within a sample, rather than modeling absolute values. For instance, applying the same shift operation to the $Q$-values of all steps within a sample still yields the same optimal ordering. As a result, the reward magnitudes across different samples do not share a consistent meaning. As stated in Corollary D.1 of the PQM paper, PQM allows comparison only when two trajectories share the same correct prefix. The limitation is validated by its suboptimal performance in our Best-of-N and beam search experiments (Section \ref{sec:BoN}, Section \ref{sec:Beam Search}), further demonstrating PQM’s deficiency in cross-sample comparability.

\paragraph{IPRM's lack of modeling for relationships between reasoning steps leads to poor cross-sample comparability.}
A core deficiency of IPRM \citep{yuan2024free} is that its training relies solely on the final outcome, which leads to ambiguous credit assignment, as the model only ensures that the aggregate of all step-level rewards aligns with the final result, without enforcing more rigorous constraints on the process. Consequently, it lacks explicit modeling of the fine-grained causal relationships between reasoning steps. The process reward for any given step is not necessarily coherent with the preceding trajectory, resulting in inconsistent process rewards that cannot be reliably compared across different samples.

\begin{figure*}[htbp]
    \centering
    \includegraphics[width=\textwidth]{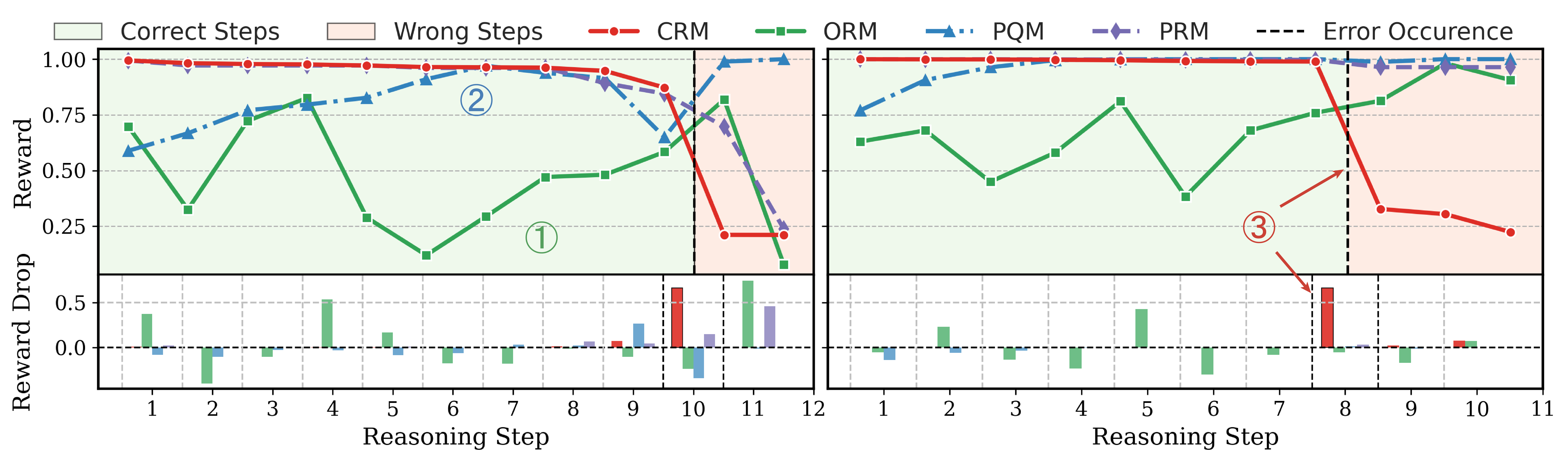}
    \caption{Case studies of different reward model scoring on trajectories from the MathShepherd validation set.}
    \label{fig:comparison}
\end{figure*}

\section{Case Studies on Process Rewards}
Figure \ref{fig:comparison} presents two case studies that compare the scoring behavior of different reward models on reasoning trajectories from the MathShepherd validation set. Each case visualizes the predicted reward per step (top) and reward drop between consecutive steps (bottom). The background is shaded to distinguish correct (green) and incorrect (red) reasoning steps, with a dashed line marking the error occurrence.

\paragraph{Our CRM precisely identifies reasoning errors.} \circledtext{1} The ORM fails to provide meaningful process rewards. Its rewards are volatile and poorly aligned with the correctness of intermediate steps. \circledtext{2} The PQM, while capturing local ranking dependencies between reasoning steps, receives only indirect supervision from the final outcome. Consequently, it may incorrectly assign high rewards after a fatal error, failing to provide a precise assessment of the final result.
\circledtext{3} In contrast, our CRM overcomes these issues and precisely identifies which reasoning step enters an incorrect state. It maintains high rewards for correct steps, exhibits an immediate drop when the first error occurs, and subsequently maintains a low reward for all following incorrect steps.

\section{Ablation Study on Process Reward Formulation of \method}
\label{appendix:process_reward_formulation}
\begin{table}[ht]
\centering
\caption{Pass@1 accuracy across benchmarks for ablation variants.}
\label{tab:log_results}
\setlength{\tabcolsep}{4pt}
\renewcommand{\arraystretch}{1.1}
\begin{tabular}{lcccccc}
\toprule
Method & MATH500 & Minerva Math & OlympiadBench & AIME25 & AIME24 & AMC23 \\
\midrule
Log-PRM & 71.4 & 33.0 & 32.2 & 13.3 & 16.6 & 45.0 \\
Log-PQM & 71.4 & 33.8 & 34.1 & 13.3 & 20.0 & 45.0 \\
Linear-\method  & 71.4 & 33.0 & 32.2 & 13.3 & 13.3 & 57.5 \\
\method    & \textbf{77.8} & \textbf{40.0} & \textbf{39.3} & \textbf{23.3} & \textbf{43.3} & \textbf{67.5} \\
\bottomrule
\end{tabular}
\end{table}

\begin{figure*}[t]
    \centering
    \includegraphics[width=\textwidth]{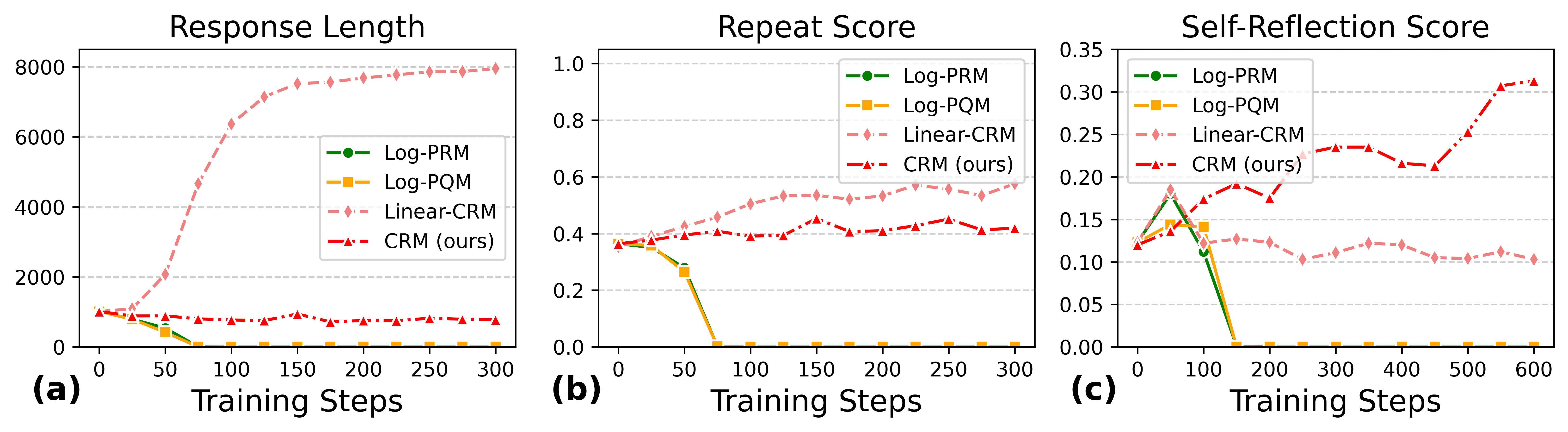}
    \caption{Training dynamics of ablation variants: (a) Response Length, (b) Repeat Score, and (c) Self-Reflection Score over training steps.}
    \label{fig:log_reward}
\end{figure*}

Our process reward is defined as $r = \log (1 - h)$, derived from the theoretical analysis in Section~\ref{sec:Modeling}.
To validate the necessity of this formulation, we conduct experiments from two perspectives: 
(1) apply a log transform to baselines (PRM and PQM) score, i.e., redefining their original per-step reward $r_t^{o}$ as $r_t = \log r_t^{o}$, denoted as Log-PRM and Log-PQM; 
and (2) further test a linearized variant of \method that replaces the original reward with $r = 1 - h$; we refer to this as Linear-\method.
Figure~\ref{fig:log_reward} shows that Log-PRM/Log-PQM and Linear-\method quickly exhibit reward hacking where response length collapses and self-reflection vanishes, whereas \method remains stable and steadily increases self-reflection.
Table~\ref{tab:log_results} mirrors this trend: \method attains the best accuracy on every benchmark with clear margins over all ablations.
These results underscore the necessity of our theoretically grounded shaping $r = \log (1 - h)$, as it offers a stable and informative process reward that effectively mitigates reward hacking.

\section{Comparison with Reasoning-enhanced PRMs}
\begin{wraptable}[10]{r}{0.5\textwidth}
\centering
\caption{Best-of-N accuracy comparison with reasoning-enhanced PRMs on MATH500.}
\resizebox{0.40\textwidth}{!}{
  \begin{tabular}{lcc}
        \toprule
        \textbf{Method} & \textbf{Best-of-8} & \textbf{Best-of-16} \\
        \midrule
        R-PRM & 49.8 & 51.2 \\
        GenPRM & 50.8 & 55.2 \\
        \midrule
        \textbf{CRM (ours)} & \textbf{53.0} & \textbf{56.4} \\
        \bottomrule
    \end{tabular}
}
\label{tab:RPRM}
\end{wraptable}
Recent work \citep{chen2025rm, guo2025reward, she-etal-2025-r, zhao2025genprm} integrates explicit reasoning capabilities into reward modeling.
To compare our \method with this line of work, we evaluate it against the representative reasoning‑enhanced reward model R‑PRM \citep{she-etal-2025-r} and GenPRM \citep{zhao2025genprm}.

We evaluated these reward models using their official released checkpoints, the same settings and inference pipelines to score identical responses from our generator on the MATH500 dataset under Best-of-N sampling. 
The results are shown in Table \ref{tab:RPRM}.
Compared to both baselines, \method consistently achieves the highest performance. Specifically, under the Best-of-8 setting, \method achieves 53.0, surpassing R-PRM (49.8) and GenPRM (50.8). Similarly, under the Best-of-16 setting, \method reaches 56.4, outperforming both R-PRM (51.2) and GenPRM (55.2). These results highlight \method's stronger capability in identifying high-quality responses compared to recent reasoning-enhanced and generative approaches.

\section{Best-of-N sampling performance characteristics}
\label{appendix:BoN_analysis}
In this section, we provide a more in-depth analysis of the performance characteristics of Best-of-N sampling.
In BoN, an LLM generator first samples $N$ candidates for each query. Subsequently, a reward model selects the response with the highest predicted reward. 
Therefore, the LLM generator plays a role in shaping BoN performance.
We observe that different LLM generators lead to distinct behaviors: BoN performance may continue to improve as $N$ increases, or it may converge at a small $N$ and show no further gains.
As shown in Table \ref{tab: MistralBoN}, with a strong generator (Llama-3.1-8B-Instruct), all methods converge at small $N$ and do not improve further, whereas with a weaker generator (MetaMath-Mistral-7B), all methods exhibit increasing BoN performance as $N$ increases.
This phenomenon can be explained as follows.
(i) Weaker LLM generators have a lower probability of producing correct responses. At small $N$, they may generate no correct candidates, leaving the reward model with nothing correct to select. As $N$ increases, the sampled candidates are more likely to include correct answers, so the BoN performance starts to improve. 
(ii) Stronger LLM generators are able to produce correct responses even with small $N$. In this case, the BoN performance is not limited, leading to faster convergence.

% This phenomenon is also closely related to another metric, Pass@N,
These points are further supported by the Pass@N metric, which measures the probability that the LLM generator produces at least one correct response among the $N$ responses, serving as an upper bound on the achievable performance of BoN sampling with any reward model. 
Table \ref{tab:generator} reports the Pass@N of the two LLM generators.
The weaker generator MetaMath-Mistral-7B shows a substantial gap between Pass@8 and Pass@128, with a relatively low Pass@8. 
This indicates that when $N$ is small, the low upper bound limits BoN performance.
The stronger LLM generator LLaMA-3.1-8B-Instruct achieves high Pass@N even for small $N$, leading to early convergence.

\begin{table}[h]
    \centering
    \caption{Best-of-N sampling performance across methods using different LLM generators.}
    \label{tab: MistralBoN}
    \resizebox{0.7\textwidth}{!}{
    \begin{tabular}{l|c|ccccc}
        \toprule
        \multirow{2}*{\textbf{LLM Generator}} & \multirow{2}*{\textbf{Method}} &  \multicolumn{5}{c}{\textbf{MATH500}} \\
        & & @8 & @16 & @32 & @64 & @128 \\
        \midrule
        \multirow{4}*{MetaMath-Mistral-7B} & Pass@N & 53.0 & 62.0 & 69.6 & 75.8 & 80.4 \\
        & PRM & 34.0 & 35.8 & 35.2 & 37.4 & 38.0\\
        & PQM & 36.4 & 38.4 & 37.8 & 38.6 & 40.4\\
        & \textbf{CRM (ours)} & 35.6 & 39.0 & 39.2 & 40.2 & 41.6\\
        \midrule
        \multirow{4}*{LLaMA-3.1-8B-Instruct} & Pass@N & 73.8 & 79.2 & 84.6 & 87.4 & 90.2 \\
        & PRM & 54.2 & 55.2 & 55.2 & 54.2 & 54.6\\
        & PQM & 53.2 & 54.4 & 54.8 & 54.8 & 55.8\\
        & \textbf{CRM (ours)} & 53.0 & 56.4 & 56.6 & 55.8 & 56.6\\
        \bottomrule
    \end{tabular}
    }
\end{table}

\begin{table}[h]
    \centering
    \caption{Pass@N of different LLM generators on MATH500 and GSM-Plus datasets.}
    \label{tab:generator}
    \resizebox{\textwidth}{!}{
    \begin{tabular}{l|cccccc|cccccc}
        \toprule
        \multirow{2}*{\textbf{LLM Generator}}  & \multicolumn{6}{c|}{\textbf{MATH500} (Pass@N)} & \multicolumn{6}{c}{\textbf{GSM-Plus} (Pass@N)} \\
        & $\Delta$[@8$\to$@128] & @8 & @16 & @32 & @64 & @128 & $\Delta$[@8$\to$@128] & @8 & @16 & @32 & @64 & @128\\
        \midrule
        MetaMath-Mistral-7B & \textbf{27.4} & 53.0 & 62.0 & 69.6 & 75.8 & 80.4 & \textbf{11.1} & 73.2 & 78.1 & 80.5 & 83.0 & 84.3 \\
        % \midrule
        LLaMA-3.1-8B-Instruct & \textbf{16.4} & 73.8 & 79.2 & 84.6 & 87.4 & 90.2 & \textbf{6.3} & 79.2 & 81.2 & 82.2 & 83.3 & 85.5  \\
        \bottomrule
    \end{tabular}
    }
\end{table}

\section{Data Efficiency Analysis of \method.}
\label{appendix:data_efficiency}
To further substantiate the data efficiency of \method, we added PRM for comparison. 
We refer to the data from the PRM training set, excluding the final step, as process data. 
We conducted experiments using the same data proportion as in our ablation study, except for 0\% (since PRM requires process data).
As shown in Table ~\ref{tab:appendix_ablation_Lz}, \method reaches near-optimal performance with only 50\% of the training data, whereas PRM still exhibits a noticeable performance gap under the same data budget.

The efficiency gain stems from how \method leverages the supervision at step $t=z$.
While PRM treats each step independently and learns a step-wise correctness classifier, \method instead models reasoning as a temporal progression toward the final answer. 
By framing the entire reasoning trajectory as a sequence converging to correctness, as derived in Eqs.~\ref{eq:L_S}, \ref{eq:L_W}, and \ref{eq:L_z}, \method allows the supervision signal at step $t=z$ to propagate backward to all earlier steps $t'\leq z$. 
This provides richer and more coherent training signals and enables \method to make more effective use of available data, resulting in significantly improved data efficiency.

\begin{table}[h]
    \centering
    \caption{Best-of-N accuracy on Math500 using varying proportions of data for \method's $\mathcal{L}_z$ and PRM.}
    \label{tab:appendix_ablation_Lz} 
    \resizebox{\textwidth}{!}{%
    \begin{tabular}{c | ccccc | c | ccccc}
        \toprule
        \multirow{2}{*}{\shortstack{Proportion of \\ data used for $L\_z$}} & \multicolumn{5}{c|}{\method} & \multirow{2}{*}{\shortstack{Process data \\ used for PRM}} & \multicolumn{5}{c}{PRM} \\
        \cmidrule{2-6} \cmidrule{8-12}
         & @8 & @16 & @32 & @64 & @128 & & @8 & @16 & @32 & @64 & @128 \\
        \midrule
        0\% & 47.0 & 44.2 & 41.6 & 39.2 & 38.2 & 0\% & - & - & - & - & - \\
        10\% & 52.4 & 51.2 & 50.6 & 49.6 & 47.6 & 10\% & 48.0 & 49.4 & 48.6 & 48.0 & 47.4 \\
        25\% & 54.0 & 52.4 & 53.6 & 53.2 & 52.0 & 25\% & 53.0 & 54.0 & 54.6 & 52.6 & 50.8 \\
        50\% & 54.4 & 53.6 & 57.2 & 55.8 & 55.0 & 50\% & 52.4 & 54.2 & 55.2 & 53.4 & 52.0 \\
        100\% & 53.0 & 56.4 & 56.6 & 55.8 & 56.6 & 100\% & 54.2 & 55.2 & 55.2 & 54.2 & 54.6 \\
        \bottomrule
    \end{tabular}
    }
\end{table}

\end{document}

%% file: math_commands.tex
\usepackage{amsmath,amsfonts,bm}

\def\eqref#1{equation~\ref{#1}}
\def\1{\bm{1}}

\DeclareMathAlphabet{\mathsfit}{\encodingdefault}{\sfdefault}{m}{sl}
\SetMathAlphabet{\mathsfit}{bold}{\encodingdefault}{\sfdefault}{bx}{n}